\documentclass[preprint,12pt]{elsarticle}

\makeatletter
\def\ps@pprintTitle{%
 \let\@oddhead\@empty
 \let\@evenhead\@empty
 \let\@oddfoot\@empty
 \let\@evenfoot\@empty
}
\makeatother

\usepackage{amssymb}

\usepackage{amsmath}

\usepackage{subcaption}
\usepackage{booktabs}

\usepackage{color}
\definecolor{red}{rgb}{1,0,0}
\definecolor{blue}{rgb}{0,0,1}
\usepackage{ulem}

\begin{document}

\begin{frontmatter}



\title{LDPoly: Latent Diffusion for Polygonal Road Outline Extraction in Large-Scale Topographic Mapping}


\author[inst1]{Weiqin Jiao\corref{cor1}}

\affiliation[inst1]{organization={Department of Earth Observation Science, Faculty of Geo-Information Science and Earth Observation (ITC), University of Twente}, 
            city={Enschede},
            postcode={7522 NH}, 
            country={The Netherlands}
            }
\cortext[cor1]{Corresponding author. Email: w.jiao@utwente.nl}
\author[inst1]{Hao Cheng}
\author[inst1]{George Vosselman}
\author[inst1]{Claudio Persello}

\begin{abstract}
Polygonal road outline extraction from high-resolution aerial images is an important task in large-scale topographic mapping, where roads are represented as vectorized polygons, capturing essential geometric features with minimal vertex redundancy. 
Despite its importance, no existing method has been explicitly designed for this task. 
While polygonal building outline extraction has been extensively studied, the unique characteristics of roads, such as branching structures and topological connectivity, pose challenges to these methods. 
To address this gap, we introduce LDPoly, the first dedicated framework for extracting polygonal road outlines from high-resolution aerial images. 
Our method leverages a novel Dual-Latent Diffusion Model with a Channel-Embedded Fusion Module, enabling the model to simultaneously generate road masks and vertex heatmaps. 
A tailored polygonization method is then applied to obtain accurate vectorized road polygons with minimal vertex redundancy. 
We evaluate LDPoly on a new benchmark dataset, Map2ImLas, which contains detailed polygonal annotations for various topographic objects in several Dutch regions. 
Our experiments include both in-region and cross-region evaluations, with the latter designed to assess the model's generalization performance on unseen regions. 
Quantitative and qualitative results demonstrate that LDPoly outperforms state-of-the-art polygon extraction methods across various metrics, including pixel-level coverage, vertex efficiency, polygon regularity, and road connectivity. 
We also design two new metrics to assess polygon simplicity and boundary smoothness. 
Moreover, this work represents the first application of diffusion models for extracting precise vectorized object outlines without redundant vertices from remote-sensing imagery, paving the way for future advancements in this field. 
\end{abstract}




\end{frontmatter}


\section{Introduction}
\label{sec:intro}
Large-scale topographic maps are fundamental cartographic products that represent the Earth's surface with high geometric accuracy and structured semantic information \cite{kent2009topographic, elberink2010acquisition, hohle2017generating}. 
They serve as critical resources for a wide range of applications, such as urban planning, infrastructure development, resource management, and military operations \cite{kent2009topographic, elberink2010acquisition}. Topographic maps encode various geographic features, such as roads, buildings, and water bodies, using structured vector representations \cite{elberink2010acquisition, hohle2017generating}. 
Among these, roads are particularly significant due to their role in autonomous driving, road network planning and mapping, traffic management, and smart city development \cite{chen2022road}. 
In topographic maps, roads are represented as closed polygons with well-defined topological structures and precisely placed vertices \cite{elberink2010acquisition}. 
An accurate and compact polygonal representation, with minimal vertex redundancy and preserved topological correctness, is essential for supporting various downstream tasks. 

\begin{figure}[tb]
  \centering
  \includegraphics[width=1.0\columnwidth]{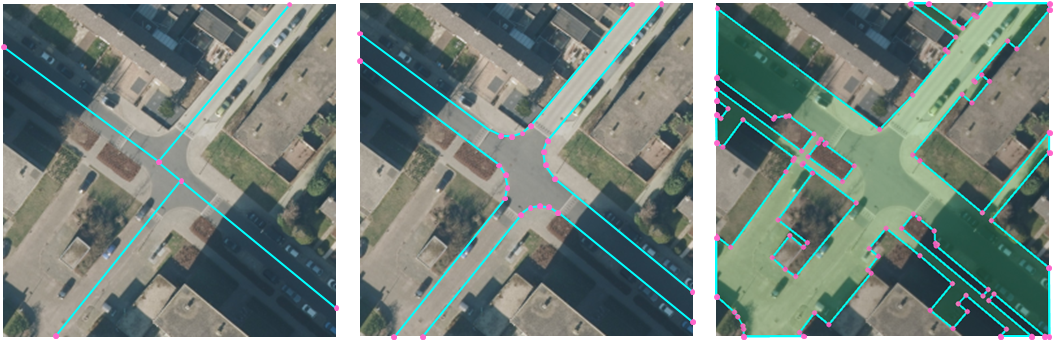}
  \caption{The differences between road networks (left), road boundaries (middle), and polygonal road outlines (right).}
  \label{fig:road_related_objects}
\end{figure}

Extracting polygonal road outlines is a challenging task due to the complex geometry and topology of road polygons in large-scale topographic maps, which are typically derived from high-resolution imagery. 
In this work, we define polygonal road outlines based on the specifications of the Dutch national topographic base, the Basisregistratie Grootschalige Topografie (BGT) \cite{GegevenscatalogusBGT}, which provides a comprehensive representation of urban, suburban and rural road surfaces. 
In BGT, a polygonal road outline refers to a contiguous polygonal region covering all road-related surfaces, regardless of material type (paved or unpaved), including motorways, bike paths, footpaths, roadside parking areas, alleys, service roads (short connecting segments linking main roads to adjacent properties), railways, airport runways and bridges. 
Moreover, the road polygons contain internal holes representing non-road elements embedded within the road network, including green spaces and traffic islands, further complicating the shape and topology of road polygons. 
These factors make manual delineation of road polygons time-consuming, making robust automated solutions essential. However, existing road extraction methods primarily focus on road masks \cite{buslaev2018fully, chaurasia2017linknet, zhou2018d, mei2021coanet, yang2022transroadnet, zhou2020bt, chen2021transunet, jiang2022roadformer, chen2023dpenet, liu2024adaptive}, road networks \cite{mattyus2017deeproadmapper, batra2019improved, li2019topological, bastani2018roadtracer, he2020sat2graph, xu2022rngdet, xu2023rngdet++}, and road boundaries \cite{xu2021icurb, xu2021topo, hu2023polyroad}, while polygonal road outlines remain unexplored. 
In road network extraction tasks, road networks are represented as graph structures, where roads are modeled using centerlines. 
Road boundary extraction methods extract road boundaries as polylines, without providing the polygonal representations of complete road regions. 
In addition, the range of road-related elements included in road boundary extraction is much narrower than polygonal road outlines, which aim to capture complete and semantically coherent road regions. 
For example, in Topo-Boundary \cite{xu2021topo}, the only publicly available dataset for road boundary detection in bird's-eye view images, defines road boundaries as individual polylines that cover only road-surface edges, airport runways, and alleys. 
It also removes certain types of scenes, such as alleys connected to the main road and overly complex intersections. 
Furthermore, these extracted road boundary instances are independent of each other, lacking spatial and semantic associations. 
Hence, Topo-Boundary is not suitable for our experiments focusing on polygonal road outline extraction. 
Figure \ref{fig:road_related_objects} illustrates the differences between these road-related objects. 

The most similar task to our work is polygonal building outline extraction \cite{zhao2018building, girard2021polygonal, xu2023hisup, castrejon2017annotating, acuna2018efficient, ling2019fast, li2019topological, zhao2021building, zorzi2022polyworld, jiao2024polyr, hu2023polybuilding, jiao2024roipoly}, which aims to extract buildings as closed polygons with potential internal holes. 
While existing methods achieve impressive results in relatively simple scenes with well-structured buildings, their performance drops significantly in complex scenarios with irregular building geometries, occlusions, and adjacent structures \cite{xu2023hisup}. 
This suggests that existing polygonal building outline extraction methods may struggle when applied to roads that exhibit complex shapes and topological structures. 
Additionally, road segmentation methods can be adapted for our task by utilizing vectorization algorithms like Douglas-Peucker \cite{douglas1973algorithms} as post-processing to generate polygonal outlines from road segmentation masks. 
However, the road masks often incur irregular boundaries \cite{xu2023hisup} and poor connectivity \cite{mei2021coanet}, leading to vectorized outputs with distorted shape and redundant vertices. 
These issues may become pronounced when dealing with complex polygonal road outlines in topographic maps. 

Furthermore, most existing methods rely on discriminative models. 
Although discriminative models excel at distinguishing foreground objects from backgrounds, they typically do not explicitly learn structured dependencies or topological relationships within objects. Consequently, this can lead to irregular road boundaries, fragmented regions, and poor connectivity, especially in scenarios with ambiguous image cues. 
To address these challenges, we propose leveraging generative models. 
Unlike discriminative models, generative models learn the underlying data distribution and generate outputs from the learned distribution, facilitating stronger reasoning capabilities regarding shape regularity and topological correctness. 
Earlier work has explored Generative Adversarial Network (GAN) \cite{zorzi2021machine} to enhance building polygon regularity. 
However, GANs are challenging to train and often encounter the model collapse problem, while further research using generative models has stagnated. 
Expanding beyond the domain of remote sensing, we observe that diffusion models have demonstrated remarkable capabilities in generating high-fidelity and highly detailed structures across various fields \cite{ho2020denoising, nichol2021improved, amit2021segdiff, wu2024medsegdiff2, rombach2022high, lin2024stable, feng2023diffpose}. 
This motivates us to leverage diffusion models for polygonal road outline extraction, addressing the limitations of existing methods in handling objects with complex geometries and topologies. 

However, directly applying a vanilla diffusion model to generate road polygons in vector format poses significant challenges. 
First, road polygons vary in vertex number, and their hierarchical structure between internal holes and external boundaries further complicates the vectorized representation. 
To overcome this obstacle, we propose a two-step pipeline. At first, a diffusion model conditioned on aerial images generates road masks and vertex heatmaps simultaneously. Then these outputs are processed via a polygonization method, which leverages the contour information of road masks and the vertex information provided by vertex heatmaps to generate vectorized road polygons with minimal vertex redundancy. 
Unlike existing polygonization methods \cite{xu2023hisup}, which overly rely on precise vertex extraction and often suffer from self-intersections, our polygonization method utilizes Gaussian heatmaps instead of sparse vertex masks \cite{xu2023hisup} to improve vertex prediction robustness, and reorganizes the vectorization logic to ensure topological consistency. 
Specifically, the final polygon vertices are selected along the predicted mask contour, guided by the extracted vertices and ordered according to their positions on the contour to naturally avoid self-intersections. 
Lastly, we incorporate the Douglas-Peucker algorithm to retain critical corners of narrow alleys that might be missed in vertex prediction while preserved by road masks, thus preventing significant shape distortions caused by the omission of a few key vertices. 

Moreover, using diffusion models in the pixel space at full image resolution is computationally expensive. 
To address this, we introduce a conditional latent diffusion model \cite{rombach2022high}, where the vertex heatmaps, road masks, and conditional inputs are mapped into a lower-resolution latent space before undergoing the diffusion-denoising process. 
This design significantly reduces computational costs and mitigates inefficiencies in optimization \cite{rombach2022high}. 
Furthermore, standard diffusion models typically involve only a single denoising objective (i.e., a clean image in image generation tasks), whereas our model involves dual denoising objectives (i.e., the latent vectors of road mask and vertex heatmap). 
This dual-target design is motivated by the nature of our two-step pipeline, where both the road mask and vertex heatmap serve as essential and complementary inputs for the subsequent polygonization process. 
Jointly modelling these two targets allows both to benefit from the generative capabilities of diffusion models within a unified generation framework. 
To ensure effective interaction between the two denoising objectives and their conditioning information, we propose a Channel-Embedded Fusion Module. 
Inspired by ChannelViT \cite{bao2023channel}, which treats hyperspectral image channels as independent semantic entities, we design learnable channel embeddings to distinguish and enhance the unique characteristics of each latent vector. 
This enables more effective feature integration and improves the generation of topologically coherent road polygons. 
We name our model LDPoly, representing its dual-latent diffusion design tailored for polygonal road outline extraction. The term ``dual" underscores the simultaneous denoising of two distinct latent targets within the diffusion framework.  

To comprehensively evaluate the effectiveness of our proposed method, we revisit existing evaluation metrics, identify their limitations, and propose two new metrics specifically designed for polygonal road outline extraction. 
In past works on road extraction and polygonal building outline extraction, various metrics have been introduced to evaluate pixel-level coverage, polygon regularity, vertex redundancy, and road connectivity. 
However, we observe that vertex redundancy is not always an ideal indicator of polygon quality, as it depends heavily on the ground truth number of vertices. For instance, ground truth annotations often densely sample curved boundaries to preserve geometric fidelity, which may hinder downstream geographic processing. In such cases, fewer vertices could sufficiently approximate the shape while maintaining usability. 
To address this limitation, we propose a new perspective, polygon simplicity, and introduce a metric called Simplicity-aware Intersection over Union (S-IoU). S-IoU rewards polygons that achieve higher IoU with fewer vertices, and, unlike redundancy-based metrics, it does not rely on ground truth vertex counts, thereby improving its generalizability. 
Moreover, unlike buildings, roads often feature long, smooth boundaries. To assess edge smoothness and penalize jagged artifacts, we introduce the Smoothness Consistency Ratio (SCR), defined as the ratio between the number of inflection points in predicted and ground truth polygons. An SCR close to 1 indicates smooth boundaries consistent with the ground truth, whereas higher values reflect irregularities and undesired edge noise.

We train and evaluate our method on a Dutch topographic map dataset, Map2ImLas, which will be publicly released as a part of an upcoming publication. 
This dataset provides polygonal outlines of various topographic classes in large-scale topographic maps of several regions in the Netherlands, including buildings, roads, vegetation, water bodies, and so on. 
We focus on the Dutch regions, as Map2ImLas is currently the only publicly available dataset that provides detailed polygonal annotations of roads in large-scale topographic maps, which can be directly used for supervised learning without additional preprocessing. 
Moreover, the Netherlands features a highly complex and well-structured road system, representative of developed regions with diverse urban, suburban, and rural topographies. These characteristics make Map2ImLas a unique and valuable resource for benchmarking polygonal road outline extraction methods. 
Specifically, we use the Deventer region as the training set and conduct evaluations on both the Deventer region and two unseen test regions: Enschede and Giethoorn. 
Deventer and Enschede cover urban and suburban areas, while Giethoorn represents rural regions, encompassing diverse road structures and topographic patterns. 
Experimental results demonstrate that our method significantly outperforms existing polygonal object outline extraction approaches across multiple evaluation metrics, and good generalization performance on unseen images. 

The main contributions are summarized as follows:
\begin{itemize}
\item A novel diffusion-model based framework, LDPoly, is proposed for polygonal road outline extraction. LDPoly employs a dual-latent diffusion model, simultaneously diffusing and denoising road masks and keypoint heatmaps, which are then vectorized into polygonal road outlines. 
To the best of our knowledge, this is the first dedicated framework for polygonal road outline extraction from high-resolution aerial imagery, and the first diffusion-based approach for generating polygonal object outlines with minimal vertex redundancy in the remote sensing domain. 
\item A Channel-Embedded Fusion Module is introduced to enhance feature interaction between the two denoising targets, road masks and vertex heatmaps, contributing to the overall improvement in extracted road polygons. 
\item We propose two new evaluation metrics to assess polygon simplicity and boundary smoothness, enabling a more comprehensive evaluation of polygonal road outlines. 
\item Experimental results compared with seven baseline methods show that our method surpasses existing methods across multiple evaluation metrics, setting a new benchmark for polygonal road outline extraction from high-resolution aerial imagery.
\end{itemize}

\section{Related work}
\label{sec:related_work}
\noindent\textbf{Road semantic segmentation.}
Road segmentation aims to extract road segmentation masks from satellite or aerial imagery. 
Over the past decade, deep learning-based road segmentation methods have been extensively studied. 
In early developments, Convolutional Neural Networks (CNNs) played a dominant role. 
Classical architectures like FCN \cite{buslaev2018fully}, SegNet \cite{badrinarayanan2017segnet}, DeepLabv3 \cite{yurtkulu2019semantic}, UNet \cite{ronneberger2015u}, LinkNet \cite{chaurasia2017linknet} were widely adopted for road extraction and later became baselines for subsequent research. 
Building upon these models, later methods focused on accurately capturing long road structures and ensuring topological integrity, thereby addressing discontinuities caused by noise and occlusions \cite{zhou2020bt}. 
For example, D-LinkNet \cite{zhou2018d} enhances LinkNet by integrating dilated convolutions, allowing for a larger receptive field without compromising spatial resolution. 
BT-RoadNet \cite{zhou2020bt} proposes a coarse-to-fine architecture that integrates boundary and topological cues via dedicated modules to solve the discontinuity issues of predicted road masks. 
CoANet \cite{mei2021coanet} leverages a strip convolution module with four directional strip convolutions, effectively capturing long-range dependencies aligned with the elongated shape of roads. 
TransRoadNet \cite{yang2022transroadnet} improves CNN-based road extraction by incorporating high-level semantic features and foreground contextual information to address CNNs' limitations in capturing global context. 
However, these CNN-based methods still have room for improvement in capturing long-range dependencies, as CNNs suffer from local receptive fields inherently. This limitation has led to a shift towards Transformer-based architectures \cite{chen2021transunet, liu2021swin}, which better capture global context and enhance road continuity. 
For example, RoadFormer \cite{jiang2022roadformer} employs a pyramidal deformable vision transformer, which adaptively focuses on the most relevant global features, effectively distinguishing roads from surrounding land covers with good topological correctness. 
DPENet \cite{chen2023dpenet} integrates CNN and Transformer architectures in a dual-path object extraction network, enabling simultaneous segmentation of buildings and roads. 
Beyond Transformer-based approaches, Fourier neural operators offer an alternative mechanism for capturing long-range dependencies and relevant features, particularly in scenarios where roads blend with similar background objects. 
For example, AFCNet \cite{liu2024adaptive} introduces an adaptive Fourier convolution network that operates in the spatial–spectral domain, effectively distinguishing roads from similar surroundings. 

However, road segmentation methods are limited to generating rasterized road masks and cannot directly produce vectorized road polygons. Furthermore, most of them focus on datasets with relatively low spatial resolution, such as the Massachusetts Roads Dataset \cite{mnih2013machine} (1.2 meter per pixel). As a result, fine road structures, such as green belts, traffic islands, and branching alleys, are not well preserved, which is not suitable for topographic map generation, as it requires high detail. 

\noindent\textbf{Road-related line-shaped object detection.}
Besides road semantic segmentation, another hot topic for roads is line-shaped object detection, including road network, road lane and road boundary extraction. 
Road networks refer to road centerlines, while road lanes denote the markings that delineate driving lanes. 
Both of them are represented as polylines. 
Methods in this domain can be broadly categorized into segmentation-based and graph-based methods. 
Segmentation-based methods generate road segmentation masks where roads are represented as thin-line structures, followed by post-processing techniques to extract vectorized representations. 
For example, DeepRoadMapper \cite{mattyus2017deeproadmapper} employed a ResNet variant to generate road segmentation masks and proposed a novel algorithm to address missing connections in the extracted road network. 
\cite{batra2019improved} proposed to use an orientation map to enhance road semantic segmentation via a multi-branch convolutional module, followed by a connectivity refinement approach to improve the extracted road networks. 
Graph-based methods, in contrast, directly generate vectorized outputs. 
Most methods employ an iterative pipeline that grows the road graph vertex by vertex. 
For instance, RoadTracer \cite{bastani2018roadtracer} used a CNN-based decision function to guide the iterative construction of the road network. 
Subsequent works such as RNGDet and its variants \cite{xu2022rngdet, xu2023rngdet++} used transformers and imitation learning to generate road network graphs iteratively. 
Other methods bypass iteration by directly predicting the road-network graph. 
Sat2Graph \cite{he2020sat2graph} proposed a novel graph-tensor encoding (GTE) approach that transforms the road graph into a tensor format, thereby eliminating the need for iterative construction. 
Recently, road boundary detection from remote-sensing imagery has attracted growing interest. 
Boundaries aligning with road-surface edges are represented as polylines. 
Most methods \cite{xu2021icurb, xu2021topo} leverage imitation learning to predict road boundaries in a graph format. 
A recent method, PolyRoad \cite{hu2023polyroad}, employed a transformer to directly predict road boundaries as vertex sequences. Notably, while these line-shaped representations effectively delineate road geometries, they are disconnected and independent, failing to provide spatial semantic information. 
In addition to remote-sensing-based road boundary detection methods, another line of research \cite{liao2022maptr, liao2024maptrv2, ding2023pivotnet} focuses on HD map reconstruction using onboard sensors such as cameras or LiDAR, primarily within the autonomous driving domain. These methods extract road boundaries, lane dividers, and pedestrian crossings as polylines or polygons \cite{hu2023polyroad}. However, similar to remote-sensing-based approaches, they treat each extracted element as an independent instance, lacking the capacity to represent a unified, topologically consistent polygonal road region. Moreover, the number of extracted map elements in HD map reconstruction is significantly fewer than those required in our task. 

\noindent\textbf{Polygonal building outline extraction.}
Although polygonal building outline extraction appears to be the most analogous task to polygonal road outline extraction, both aiming to represent objects as closed polygons with potential internal holes, the geometric and topological characteristics of roads differ significantly from buildings, which poses challenges to the generalization between these two tasks. 
There are two broad categories of existing building polygon extraction methods: segmentation-based and end-to-end methods. 
Segmentation-based methods \cite{zhao2018building, girard2021polygonal, xu2023hisup} first employ segmentation models to generate building masks, followed by polygonization techniques to vectorize the extracted outlines. 
In contrast, end-to-end methods \cite{castrejon2017annotating, acuna2018efficient, ling2019fast, zorzi2022polyworld, li2019topological, zhao2021building, jiao2024polyr, jiao2024roipoly} directly predict building outlines as ordered vertex sequences. 
While both categories achieve strong performance on datasets with relatively simple building structures, such as CrowdAI \cite{mohanty2020deep}, their effectiveness deteriorates in more complex environments, such as the Inria dataset \cite{maggiori2017can}. 
This performance drop indicates their limitations in handling objects with intricate and irregular outlines, such as roads. 

\noindent\textbf{Diffusion models.}
Diffusion models (DMs) \cite{ho2020denoising, batra2019improved} operate by progressively adding noise to an image during a forward process and learning to reverse it step by step, effectively generating new images from noise during inference. 
Compared to discriminative models, DMs excel in modeling complex data distributions, enabling high-quality outputs with improved robustness. 
Leveraging these advantages, DMs have made significant strides in various computer vision tasks, including dense prediction problems closely related to polygonal road outline extraction, such as image segmentation \cite{baranchuk2021label, wu2024medsegdiff, wu2024medsegdiff2, lin2024stable} and keypoint heatmap prediction \cite{feng2023diffpose}. 
In image segmentation, conditional diffusion models have been employed to generate object masks from Gaussian noise given an input image as the condition \cite{amit2021segdiff, wu2024medsegdiff, wu2024medsegdiff2}. However, as they operate directly in pixel space, they incur substantial computational costs and pose optimization challenges \cite{rombach2022high, lin2024stable}. 
To address these issues, \cite{lin2024stable} proposed using Latent Diffusion Models (LDMs) \cite{rombach2022high} for medical image segmentation. Unlike conventional DMs, LDMs perform diffusion and denoising in a lower-resolution latent space, significantly reducing computational overhead. 
Beyond segmentation, DMs have also demonstrated strong performance in keypoint heatmap prediction of human pose estimation \cite{feng2023diffpose}. 
However, unlike human pose estimation with only one keypoint per image, road outline extraction involves predicting numerous keypoints with highly irregular spatial distributions, posing additional challenges.  

Despite the remarkable success of diffusion models, especially the latent diffusion paradigm, in related tasks, their application to polygonal road outline extraction remains unexplored. 
Therefore, we propose a dual latent diffusion model that jointly denoises road masks and vertex heatmaps. 
Additionally, we introduce a novel fusion module to enhance their interaction during denoising. 
This design ensures regular, smooth road masks and precisely located vertices, leading to accurate and regular polygonal road outlines without redundant vertices. 

\section{Preliminaries}
In this section, we provide a brief overview of diffusion models and latent diffusion models to lay the groundwork for our proposed method. 

\noindent\textbf{Diffusion model.}
Diffusion models \cite{ho2020denoising, batra2019improved} learn to approximate complex data distributions by gradually recovering meaningful data from a sample of Gaussian noise. 
This is achieved through a two-step process: (1) a forward diffusion process that gradually corrupts an input sample $x_0$ by adding Gaussian noise over $T$ timesteps, and (2) a reverse process that learns to reconstruct the original sample from a random Gaussian noise iteratively. 
The forward process follows a Markovian chain: 
\begin{equation}
\label{equ:foward}
    q(x_t \mid x_0) = \mathcal{N} \left( x_t \mid \sqrt{\bar{\alpha}_t} x_0, (1 - \bar{\alpha}_t) \mathbf{I} \right),
\end{equation}
where $\bar{\alpha}_t = \prod_{p=0}^{t} \alpha_p$, $\alpha_p = 1 - \beta_p$ and $\beta_p$ are predefined variance-controlled noise schedule controlling the noise injection at each step $p$. 
Building on this, the forward Markovian chain can be reformulated using the reparameterization trick as: 
\begin{equation}
\label{equ:forward_diffusion}
    x_t = \sqrt{\bar{\alpha}_t} x_0 + \sqrt{1 - \bar{\alpha}_t} \epsilon_t, \quad \epsilon_t \sim \mathcal{N}(0, \mathbf{I}).
\end{equation}
On the other hand, the posterior distribution $q(x_{t-1}|x_t, x_0)$ can be derived from Bayes' theorem as a Gaussian distribution: 
\begin{equation}
    \label{equ:posterior}
    q(x_{t-1} \mid x_t, x_0) = \mathcal{N} \left( x_{t-1} \mid \tilde{\mu}(x_t, x_0), \tilde{\beta_t}\mathbf{I} \right),
\end{equation}
where $\tilde{\beta_t}$ is a hyperparamter, and $\tilde{\mu}$ can be derived either from $x_t$ and $x_0$ or, alternatively, from $x_t$ and $\epsilon_t$ using Equation \ref{equ:forward_diffusion}. 
Consequently, in Denoising Diffusion Probabilistic Model (DDPM) \cite{ho2020denoising}, the neural network $f_{\theta}(x_t, t)$ is trained to predict either the noise $\epsilon_t$ or the original sample $x_0$ from $x_t$ given the sampling step $t$. 
In conditional diffusion models, the denoiser $f_{\theta}(x_t, t, I)$ incorporates an additional conditioning input, such as an image $I$ to guide the generation process. 
During inference, DDPM \cite{ho2020denoising} refines a noise sample $x_T$ following the posterior $q(x_{t-1}|x_t, x_0)$ step by step, from $x_T \rightarrow x_{T-1} \rightarrow x_{T-2}$, until the final output $x_0$ is reconstructed. 
To improve sampling efficiency, Denoising Diffusion Implicit Model (DDIM) \cite{batra2019improved} introduces a non-Markovian sampling strategy, which accelerates the sampling process by skipping intermediate steps in a deterministic manner. 
This makes it computationally more efficient than DDPM while preserving generation quality. 

\noindent\textbf{Latent Diffusion model.}
Despite their success, standard diffusion models operate in high-resolution pixel space, leading to high computational costs and inefficiencies. 
To address this, Latent Diffusion Models (LDMs) \cite{rombach2022high} introduce a low-resolution latent space where diffusion and denoising are performed, significantly reducing the dimensionality of the learning process. 
LDMs first transform images into a lower-resolution latent representation using the encoder $E$ of a pretrained autoencoder: 
\begin{equation}
    z = E(x).
\end{equation}
In particular, the autoencoder used in \cite{rombach2022high} consists of a convolutional encoder $E$ and a convolutional decoder $D$, which are trained separately from the diffusion model to reconstruct the input image from its latent representation. 
Its parameters are kept frozen during the training of the diffusion model. 
In LDMs, the forward process then becomes: 
\begin{equation}
    z_t = \sqrt{\bar{\alpha}_t} z_0 + \sqrt{1 - \bar{\alpha}_t} \epsilon_t, \quad \epsilon_t \sim \mathcal{N}(0, \mathbf{I}),
\end{equation}
where $z_t$ represents the noisy latent vector at timestep $t$. 
After obtaining the clean latent vector $z_0$, a pretrained decoder $D$ effectively maps it back to pixel space in a single forward pass. 

In this paper, we propose LDPoly, a dual-latent diffusion model designed to simultaneously generate road masks and vertex heatmaps. 
In our formulation, the data samples consist of road masks $m \in \mathbb{R}^{h \times w}$ and vertex heatmaps $k \in \mathbb{R}^{h \times w}$, where $h$ and $w$ are the height and width of the image.  The model $f_{\theta}(m_t, k_t, t, I)$ is trained to reconstruct $m_0$ and $k_0$ from noisy inputs $m_t$ and $k_t$, conditioned on the corresponding image $I$. During inference, the model generates $m_0$ and $k_0$ by progressively denoising an initial random Gaussian noise $\epsilon \in \mathbb{R}^{h \times w}$. 

\section{Methodology}
\begin{figure}[tb]
  \centering
\includegraphics[trim=50pt 150pt 120pt 50pt, clip, width=1.0\columnwidth]{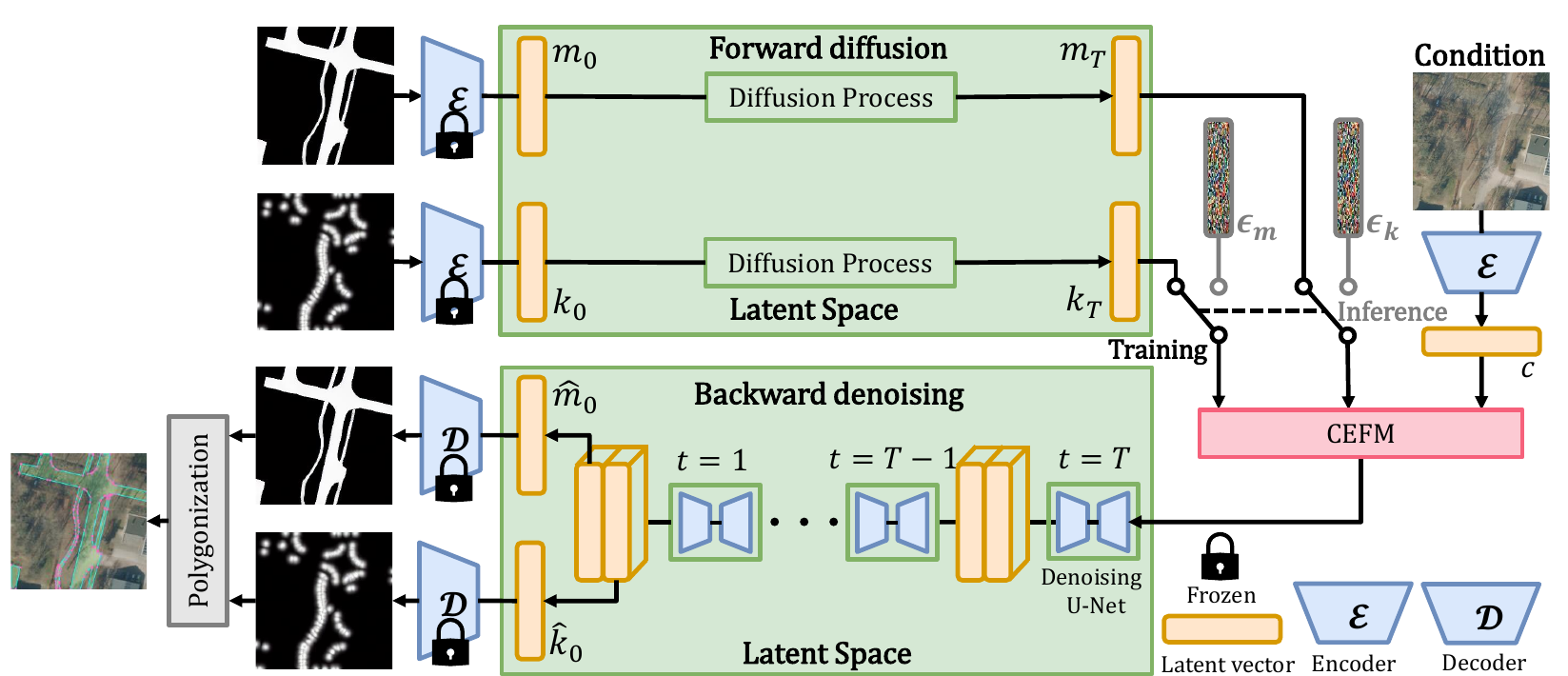}
  \caption{The overall architecture of our proposed method LDPoly. The double-added double-throw switch symbol indicates that during training, the corrupted latent vectors of the road mask and vertex heatmap are fed into the Channel Embedded Fusion Module (CEFM), whereas during inference, randomly generated Gaussian noise with the same size as the latent vectors is used as input to CEFM. The output road mask and vertex heatmap are further processed by a polygonization algorithm to extract the polygonal road outline.}
  \label{fig:network_architecture}
\end{figure}

The overall architecture of our proposed model is illustrated in Figure \ref{fig:network_architecture}. It consists of pretrained autoencoders for encoding the input into the latent space, a dual-latent diffusion process for progressively refining the latent representations, a denoising UNet, and the proposed Channel-Embedded Fusion Module. 
The model takes an input image as a condition and generates road masks and vertex heatmaps, which are then processed using our proposed polygonization algorithm to obtain vectorized road polygons. 

In the following sections, we introduce the data preprocessing pipeline, the training and inference strategies, the denoising network, and the proposed fusion module. 
Finally, we describe the proposed polygonization algorithm. 

\subsection{Data preprocessing and latent representation encoding}
\begin{figure}[tb]
  \centering
\includegraphics[trim=10pt 10pt 10pt 10pt, clip, width=0.8\columnwidth]{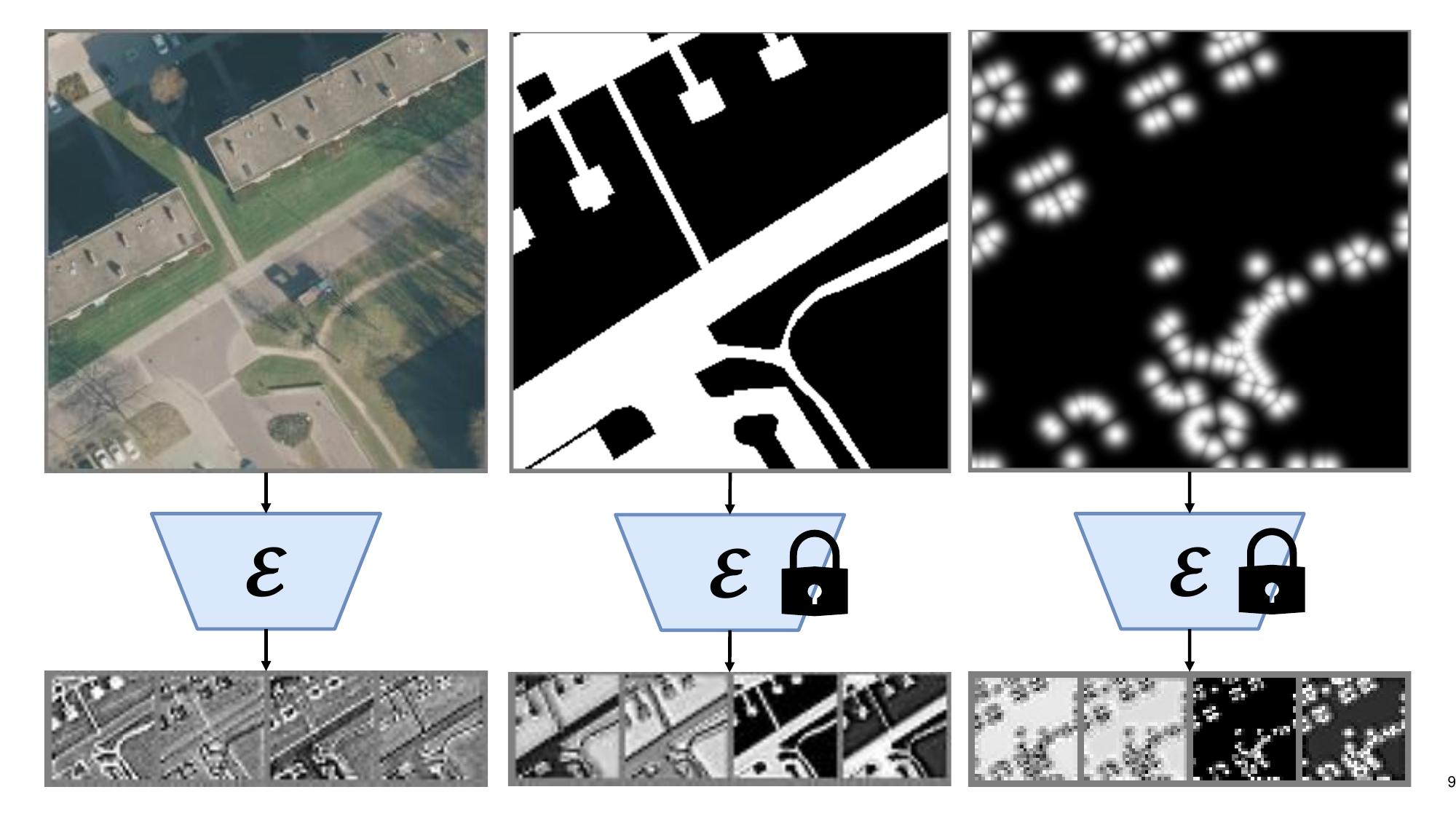}
  \caption{An example of the training input, including the aerial image (left), the road mask (middle), and the vertex heatmap (right), along with their corresponding latent representations and the encoder $\varepsilon$ of the autoencoder. The latent representations, with a last dimension of $c$, are visualized by mapping each channel separately to a grayscale image. In our experiments, $c=4$. Hence, each latent representation is displayed as a row of four grayscale images.}
\label{fig:data_encoding_pipeline}
\end{figure}

The training input and the latent representation encoding process are illustrated in Figure \ref{fig:data_encoding_pipeline}. 
As shown in the figure, the training input consists of an aerial image, a corresponding road mask, and a vertex heatmap. 
Since the dataset only provides vectorized road polygons, the ground truth road mask and vertex heatmap for training must be generated from these vector annotations. 
The road mask is generated by rasterizing the reference polygons into binary masks, while the vertex heatmap is constructed by representing each vertex as a localized 2D Gaussian distribution with a standard deviation of 5 pixels. 
This value is empirically chosen to balance sparsity and density, ensuring that adjacent vertices remain distinguishable while maintaining sufficient continuity for effective vertex extraction. 

Given the ground truth road masks and vertex heatmaps, denoted as $m_0 \in \mathbb{R}^{h \times w}$ and $k_0 \in \mathbb{R}^{h \times w}$, the encoder of the pretrained autoencoder from Stable Diffusion \cite{rombach2022high} is used to obtain their latent representations: 
\begin{equation} 
z_{m_0} = E(m_0) \in \mathbb{R}^{\frac{h}{r} \times \frac{w}{r} \times c}, \quad z_{k_0} = E(k_0) \in \mathbb{R}^{\frac{h}{r} \times \frac{w}{r} \times c}, 
\end{equation}
where $r$ is the downsampling factor, and $c$ is the latent space dimensionality. 

Due to the relatively simple structure of masks and heatmaps, we observe that the pretrained decoder can already reconstruct precise road masks and vertex heatmaps from their corresponding latent vectors:
\begin{equation} 
    \hat{m}_0 = D(E(m_0)), \quad \hat{k}_0 = D(E(k_0)). 
\end{equation}
Therefore, we freeze these autoencoders during training to ensure stable feature extraction. 

Additionally, we introduce another encoder to process the conditioning image $I$, obtaining its latent representation: 
\begin{equation} 
    z_I = E(I), 
\end{equation}
Unlike the encoders for road masks and vertex heatmaps, this encoder remains trainable, allowing it to learn task-specific feature representations, following \cite{lin2024stable}. 

\subsection{Training and inference}
The training process follows a standard diffusion model paradigm. 
We gradually corrupt the latent representations $z_{m_0}$ and $z_{k_0}$ by adding Gaussian nosie at each timestep t:
\begin{equation}
    z_{m_t} = \sqrt{\bar{\alpha}_t} z_{m_0} + \sqrt{1 - \bar{\alpha}_t} \epsilon_{m_t}, \quad \epsilon_{m_t} \sim \mathcal{N}(0, \mathbf{I}),
\end{equation}
\begin{equation}
    z_{k_t} = \sqrt{\bar{\alpha}_t} z_{k_0} + \sqrt{1 - \bar{\alpha}_t} \epsilon_{k_t}, \quad \epsilon_{k_t} \sim \mathcal{N}(0, \mathbf{I}).
\end{equation}
The denoiser network $f_{\theta}(z_{m_t}, z_{k_t}, t, I)$ is then trained to predict the noise components $\hat{\epsilon}_{m_0}$ and $\hat{\epsilon}_{k_0}$. 
Leveraging the reparameterization trick, we can also recover the latent vectors from the predicted noise components:
\begin{equation}
    \hat{z}_{m_0} = \frac{z_{m_t} - \sqrt{1 - \bar{\alpha}_t} \hat{\epsilon}_{m_t}}{\sqrt{\bar{\alpha}_t}}.
\end{equation}
\begin{equation}
    \hat{z}_{k_0} = \frac{z_{k_t} - \sqrt{1 - \bar{\alpha}_t} \hat{\epsilon}_{k_t}}{\sqrt{\bar{\alpha}_t}}.
\end{equation}

The training objective contains two components. 
First, we minimize the $\mathcal{L}_1$ loss between the predicted and true noise values. 
Second, following \cite{lin2024stable}, we impose another $\mathcal{L}_1$ loss directly on the recovered latent vectors to further enhance supervision. 
The overall loss $\mathcal{L}$ is defined as:
\begin{equation}
\label{equ:loss}
    \mathcal{L} = \lambda_{m_t}\left| \epsilon_{m_t} - \hat{\epsilon}_{m_t} \right| + \lambda_{k_t}\left| \epsilon_{k_t} - \hat{\epsilon}_{k_t} \right| + \lambda_{m_0}\left| z_{m_0} - \hat{z}_{m_0} \right| + \lambda_{k_0}\left| z_{k_0} - \hat{z}_{k_0} \right|.
\end{equation}
where $\lambda_{m_t}$, $\lambda_{k_t}$, $\lambda_{m_0}$ and $\lambda_{k_0}$ are the coefficients of different loss terms. 

During inference, we initialize two randomly sampled Gaussian noise with the same shape as the latent vectors, i.e., $\frac{h}{r}\times \frac{w}{r} \times c$. 
These noise vectors, along with the latent representation of the conditioning image, are fed into the denoising network, where they undergo iterative refinement through the DDIM sampling process. 
This progressively removes noise and reconstructs the latent road mask and vertex heatmap representations. 
Finally, the pretrained decoder transforms the denoised latent representations back into pixel space, generating the final road masks and vertex heatmaps.

\subsection{Denoising network and channel-embedded fusion module}
We built the denoiser upon the UNet architecture in Stable Diffusion \cite{rombach2022high}, which adopts a hierarchical encoder-decoder structure with skip connections. 
Each layer of the encoder and decoder consists of residual blocks (ResBlocks) for feature extraction, self-attention mechanisms for capturing long-range dependencies, and timestep conditioning to incorporate diffusion step information into the network. 
To integrate conditional information, Stable Diffusion adopts two strategies: (1) Cross-attention at each UNet layer, where the conditional features serve as keys and values. (2) Concatenation of the latent representation of the conditional input and the denoising target along the last channel dimension before feeding it into the network. 

However, directly applying these fusion strategies to our scenario is suboptimal due to the fundamental differences between our task and Stable Diffusion's standard setup. 
In Stable Diffusion, the denoising network processes a single denoising target, allowing it to focus on modeling the semantic relationship between the target and the conditional input in a straightforward manner. 
In contrast, our model introduces two distinct denoising targets: 
(1) road masks: dense, binary representations capturing regional connectivity; 
(2) road polygon vertex heatmaps: grayscale representations with sparse Gaussian peaks marking keypoints along road outlines. 
This requires the network to jointly reason across multiple interactions: 
(1) between the road mask and the conditioning image, 
(2) between the vertex heatmap and the conditioning image, and (3) between the road mask and the vertex heatmap, as these heterogeneous inputs encode complementary but structurally different aspects of road geometry. 
This multi-level interaction introduces additional complexity, requiring the network to not only effectively model these relationships but also maintain clear distinctions to prevent confusion and interference during the denoising process. 
\begin{figure}[tb]
  \centering
  \includegraphics[trim=90pt 30pt 90pt 30pt, clip, width=0.8\columnwidth]{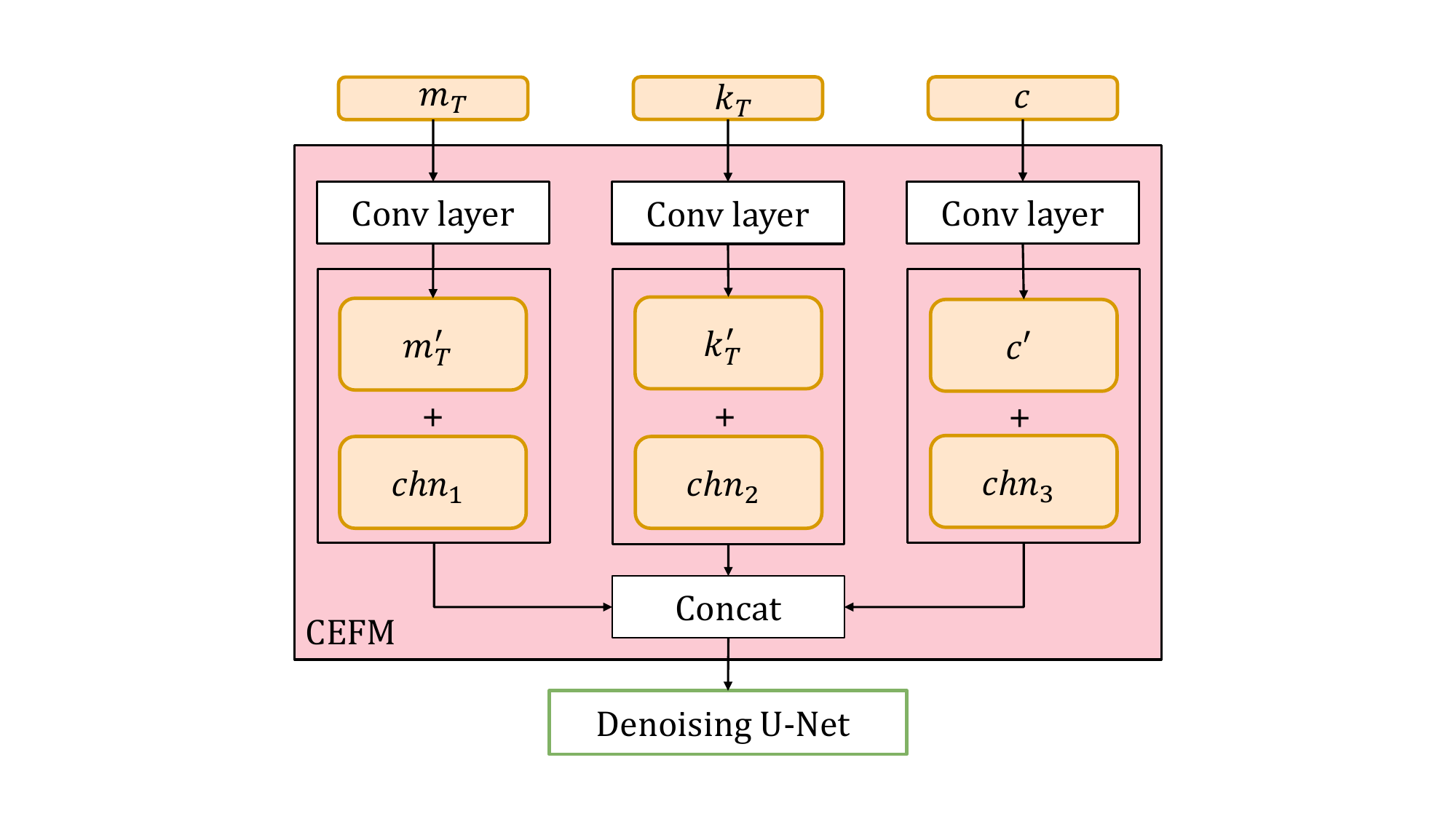}
  \caption{The detailed structure of the Channel-Embedded Fusion Module. Here, $m'_T$, $k'_T$ and $c'$ represent the high-dimensional latent features of the road mask, vertex heatmap, and conditioning image, respectively. $\text{chn}_1$, $\text{chn}_2$, and $\text{chn}_3$ are channel-specific offsets added to their corresponding latent features.}
  \label{fig:CEFM}
\end{figure}

To address this challenge, we propose the Channel-Embedded Fusion Module (CEFM), inspired by ChannelViT \cite{bao2023channel}, which enhances the semantic distinctiveness of different channels in hyperspectral images by using learnable channel embeddings. 
The detailed structure of CEFM is shown in Figure \ref{fig:CEFM}. 
In Stable Diffusion, the pretrained autoencoder produces highly compressed latent representations with only four channels and a downsampled spatial resolution, retaining only key global semantic information. 
This compression limits their expressiveness and makes it challenging to effectively fuse multiple latent representations. 
To overcome this limitation, CEFM first applies convolutional layers to project the latent representations into a higher-dimensional feature space, enriching their expressiveness. 
Next, high-dimensional learnable channel embeddings are introduced to explicitly differentiate the three high-dimensional latent features: 
\begin{equation}
    \text{chn}_i\in \mathbb{R}^{\frac{h}{r} \times \frac{w}{r} \times d}, i=1,2,3,
\end{equation}
where each embedding $\text{chn}_i$ serves as a channel-specific offset and is directly added to its respective latent feature. 
These distinguished and enhanced latent features are then concatenated along the channel dimension to enable effective fusion. 

Finally, the denoising UNet takes the concatenated latent features as input and outputs a latent vector with twice the original channel dimensionality $c$, which is then split along the last dimension to obtain separate predictions for $\hat{\epsilon}_{m_0}$ and $\hat{\epsilon}_{k_0}$. 

\subsection{Polygonization.}
Given the reconstructed road mask $\hat{m}_0$ and vertex heatmap $\hat{k}_0$, we propose a polygonization method to extract vectorized road polygons. 
Since predicting dense masks is generally less challenging than localizing sparse vertices, the positions of the extracted vertices may be less reliable than the mask contours. 
Hence, we revise the commonly used polygonization logic \cite{xu2023hisup}, where unordered extracted vertices are connected based on the sequence information provided by the mask contour to form polygons. 
This common polygonization method leads to heavy reliance on the precision of extracted vertices and may cause self-intersections between edges due to vertex misalignment. 
In our approach, instead of directly using the extracted vertices to construct the polygon, we use them as guidance to select corresponding points from the ordered mask contour. 
This ensures that the final polygon follows the actual mask geometry and avoids self-intersections. 
Furthermore, we introduce a refinement step to recover inflection points that may be missed due to closely spaced corners in narrow road regions. 
Our complete polygonization procedure consists of four main stages: dense polygon extraction, keypoint-guided vertex selection, distance-based vertex filtering, and a refinement step to recover missing inflection points. The details of each step are provided in the following subsections. 

\noindent\textbf{Dense polygon extraction.}
We first extract an initial dense polygon candidate $C = \{v_1, v_2, ..., v_n\}$ by tracing the boundary of the predicted road mask $\hat{m}_0$. 
This results in a densely sampled polygon with an excessive number of vertices. 

\noindent\textbf{Keypoint-guided vertex selection}
To reduce redundant vertices of $C$ while preserving structural keypoints along road outlines, we leverage the predicted vertex heatmap $\hat{k}_0$. 
Unlike HiSup \cite{xu2023hisup}, which directly predicts a vertex classification mask, our method follows the keypoint heatmap prediction strategy commonly used in pose estimation \cite{feng2023diffpose}, which improves vertex localization accuracy and mitigates the sparsity issues found in segmentation-based approaches. 
As a result, by simply utilizing Non-Maximum Suppression (NMS) to $\hat{k}_0$ with a fixed threshold, we are able to extract a sparse set of vertices $V = \{v_1', v_2', ..., v_m'\}$, which already exhibit strong spatial sparsity and serve as effective reference points for polygon simplification. 

\noindent\textbf{Distance-based vertex filtering}
Given the vertex set $V$, we perform a distance-based filtering step to remove redundant vertices from the dense polygon $C$. 
For each vertex $v \in C$,  we compute its Euclidean distance to the nearest extracted keypoint $v' \in V$:
\begin{equation} d(v, V) = \min_{v' \in V} || v - v' ||_2. \end{equation}
Vertices in $C$ are retained only if their distance to the nearest keypoint is below a predefined threshold $d_\text{th}$:
\begin{equation} C' = \{ v \in C \mid d(v, V) < d_\text{th} \}. \end{equation}
Since the retained points in $C'$ are directly derived from the road mask contour, we ensure no topological inconsistency in the final polygon. 

\noindent\textbf{Retaining missing inflection points}
Although the proposed polygonization method effectively extracts consistent and topologically accurate road polygons, certain challenging cases may still lead to missing critical inflection points. 
One such scenario occurs at junctions where narrow and elongated alleys intersect perpendicularly with a main road. 
Due to the small road width, these junctions are extremely close to each other (often within 5 pixels) making them difficult to distinguish in the keypoint heatmap. 
To address this issue, we propose to take advantage of the conventional polygonization method, the Douglas-Peucker (DP) algorithm, to recover missing inflection points from the mask contour. 
Notably, directly applying the DP algorithm for polygonization is ineffective: a large tolerance threshold oversimplifies the shape, while a small threshold retains excessive redundant vertices, making it difficult to strike a balance. 
Nevertheless, we can extract inflection points from its output to help refine our results. 
We first apply the DP algorithm with a small tolerance threshold $\epsilon$ to generate an intermediate polygon $\hat{c}$,  which retains all significant shape details but also introduces significantly redundant vertices. 
We define inflection points as vertices where the angle between two adjacent edges falls within a range around 90°, indicating a road junction:
\begin{equation} 
V_{\text{inflection}} = { v \in \hat{C} \mid 90^\circ - \tau \leq \theta(v) \leq 90^\circ + \tau } 
\end{equation}
where $\theta(v)$ is the angle at vertex $v$, and $\tau$ is a predefined threshold. 
Finally, we merge these selected inflection points into the refined polygon $C'$, and sort them based on their positions in $C$ to get the final polygon. 
A detailed example of the results before and after the inflection points have been retained can be found in Figure \ref{fig:retain_inflection_point} in Section \ref{sec:ablation}. 
This refinement further avoids notable shape distortion and introduces additional gains to the pixel-level coverage of the extracted polygons, as demonstrated in Section \ref{sec:ablation}. 

\section{Experiment}
\subsection{Experimental setup}
\label{sec:experimantal_setup}
\begin{figure}[tb]
    \centering
    \includegraphics[width=0.5\textwidth]{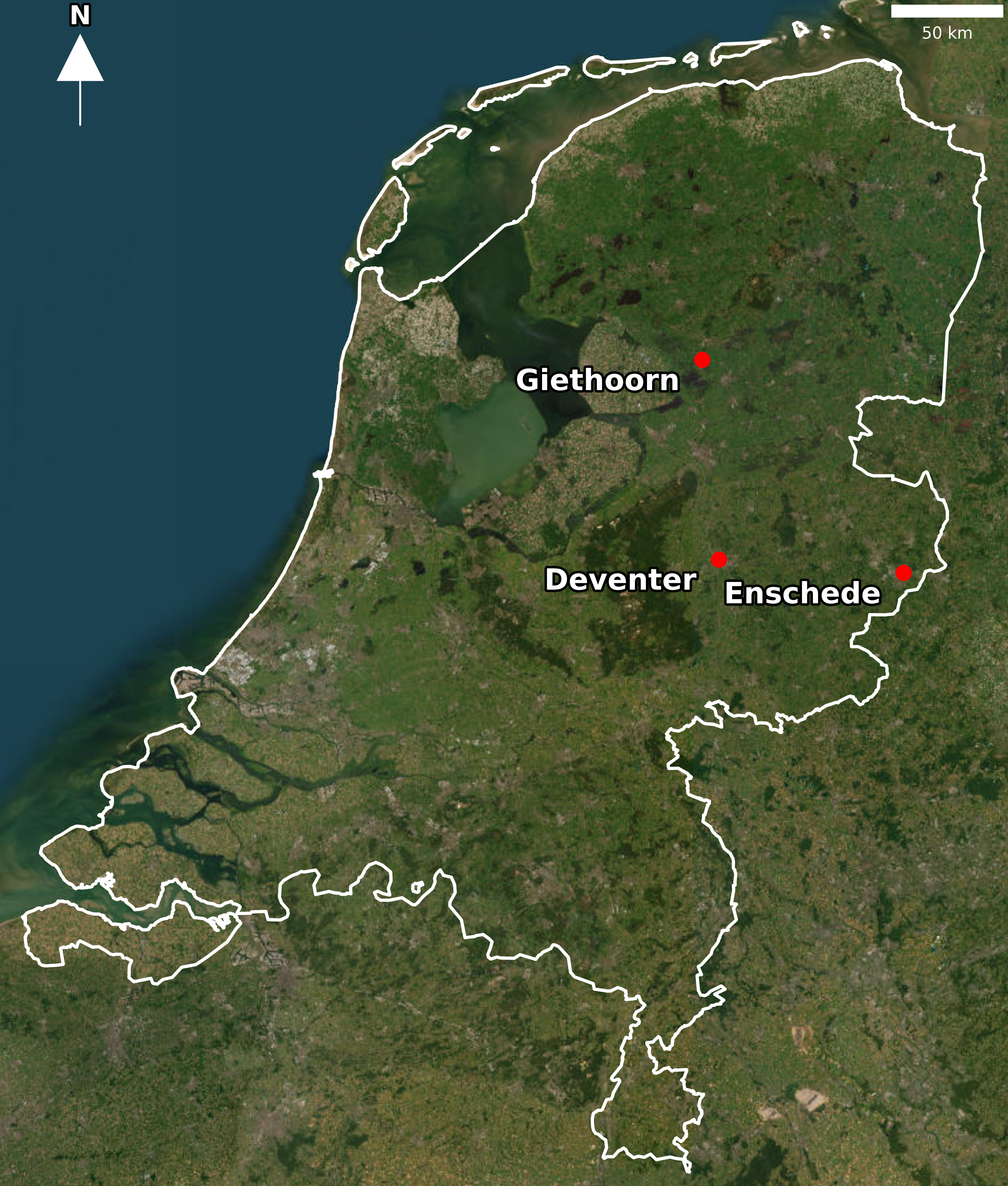}
    \caption{Study area map indicating the locations of Deventer, Enschede, and Giethoorn in the Netherlands. The base map imagery is accessed via Esri’s World Imagery service, which incorporates data from multiple sources including commercial satellites and national mapping agencies. Administrative boundaries are derived from GADM data https://gadm.org/.}
    \label{fig:study_area_map}
\end{figure}

\noindent\textbf{Dataset.} 
We evaluate LDPoly on the Dutch topographic map dataset, Map2ImLas, which includes several regions in the Netherlands, covering urban, suburban, rural and forested landscapes. 
The dataset provides polygonal annotations for various geographic objects, such as roads, buildings, water, and vegetation. 
Each RGB image has a size of 4000$\times$4000 pixels with a 7.5 cm spatial resolution. 
We use the Deventer region for both training and in-region testing, and conduct cross-region evaluations on the Enschede and Giethoorn regions. 
Both the Deventer and Enschede regions cover urban and suburban landscapes, while the Giethoorn region covers rural and forested landscapes. 
This setup enables a comprehensive evaluation of the model’s generalization capability. 
The geographical locations of the study areas are illustrated in Fig. \ref{fig:study_area_map} for reference. 
The Deventer region contains 224 images for training and 28 images for testing. 
The Enschede and Giethoorn regions include 25 and 26 test images, respectively. 
Given the high spatial resolution of the dataset, and that slightly lower resolutions remain sufficient for outline delineation, we downsample each image from 4000$\times$4000 to 1024$\times$1024. 
To reduce computational burden and fit within resource constraints, each image is then split into 16 non-overlapping patches. 

\noindent\textbf{Implementation Details.}
We adopt the KL-regularized continuous latent autoencoder from the pretrained Stable Diffusion model. 
Specifically, we use a downsampling rate of 8 and a latent dimensionality of 4. 
As a result, an input image of size 256$\times$256$\times$3 is encoded into a latent representation of size 32$\times$32$\times$4. 
For training, we use an initial learning rate of 4e-5 with a batch size of 4, and train the model for 850 epochs. 
At epoch 650, the learning rate is decayed to 4e-6. 
We also apply an early stopping strategy to prevent overfitting. 
Data augmentation is limited to random flipping and random rotation. 
The angle threshold $\tau$ for defining inflection points during polygonization is set to $30^\circ$. 
The coefficients of the different loss terms in Equation \ref{equ:loss}, $\lambda_{m_t}$, $\lambda_{k_t}$, $\lambda_{m_0}$, and $\lambda_{k_0}$, are uniformly set to 1, considering the comparable importance and magnitude of each term. 
The complete code, pretrained weights, and processed dataset will be released upon acceptance of this paper.

\subsection{Evaluation metrics} 
We evaluate our method based on the following key aspects: pixel-level coverage, vertex redundancy, polygon simplicity, polygon regularity and road connectivity. 

\noindent\textbf{Pixel-level coverage} Pixel-level coverage refers to the overall alignment between the predicted road polygon and the ground truth. 
We use Intersection-over-Union (IoU) and its variant, Boundary IoU (B-IoU) \cite{cheng2021boundary}, to measure coverage quality. 
IoU evaluates the general overlap between predicted and ground truth masks. 
B-IoU specifically measures the intersection-over-union between the dilated boundary regions (pixels within a fixed distance from the boundary) of the ground truth and predicted masks, emphasizing the accuracy of boundary alignment. 

\noindent\textbf{Vertex redundancy.} A well-formed road polygon should not only achieve high pixel-level coverage but also maintain a compact and non-redundant vertex representation, which is crucial for large-scale topographic mapping and subsequent geospatial processing. 
To quantify vertex redundancy, we employ N-Ratio \cite{zorzi2022polyworld} and Complex-IoU (C-IoU) \cite{zorzi2022polyworld}. 
N-Ratio calculates the ratio of the predicted polygon vertices to the ground truth vertices. 
C-IoU refines the standard IoU by introducing a factor that penalizes excessive or insufficient vertex counts:
\begin{equation}
    \text{C-IoU}(m, \hat{m}) = \text{IoU}(m, \hat{m}) \cdot (1 - \frac{N_m - N_{\hat{m}}}{N_m + N_{\hat{m}}})
\end{equation}
where $N_m$ and $N_{\hat{m}}$ denote the number of vertices in the ground truth and predicted polygons, respectively. $\text{IoU}(m, \hat{m})$ is the IoU between the ground truth road polygon $m$ and the predicted polygon $\hat{m}$. 

\noindent\textbf{Polygon simplicity.} However, in certain cases, a lower vertex number than the ground truth is desirable for topographic map generation. 
For instance, while curved boundaries can be represented using many vertices, a more compact polygon with fewer vertices that still captures the overall shape is often preferred in geospatial applications. 
To evaluate this aspect, we introduce the concept of polygon simplicity, which aims to quantify the trade-off between vertex density and shape preservation. 
Specifically, we propose Simplicity-aware IoU (S-IoU) to encourage models to achieve high IoU while using fewer vertices:
\begin{equation}
    \text{S-IoU}(m, \hat{m}) = \text{IoU}(m, \hat{m}) \cdot \text{SF}(N_{\hat{m}})
\end{equation}
where SF is the simplicity factor which penalizes predictions with excessive vertices. 
\begin{figure}[tb]
  \centering
\includegraphics[width=1.0
\columnwidth]{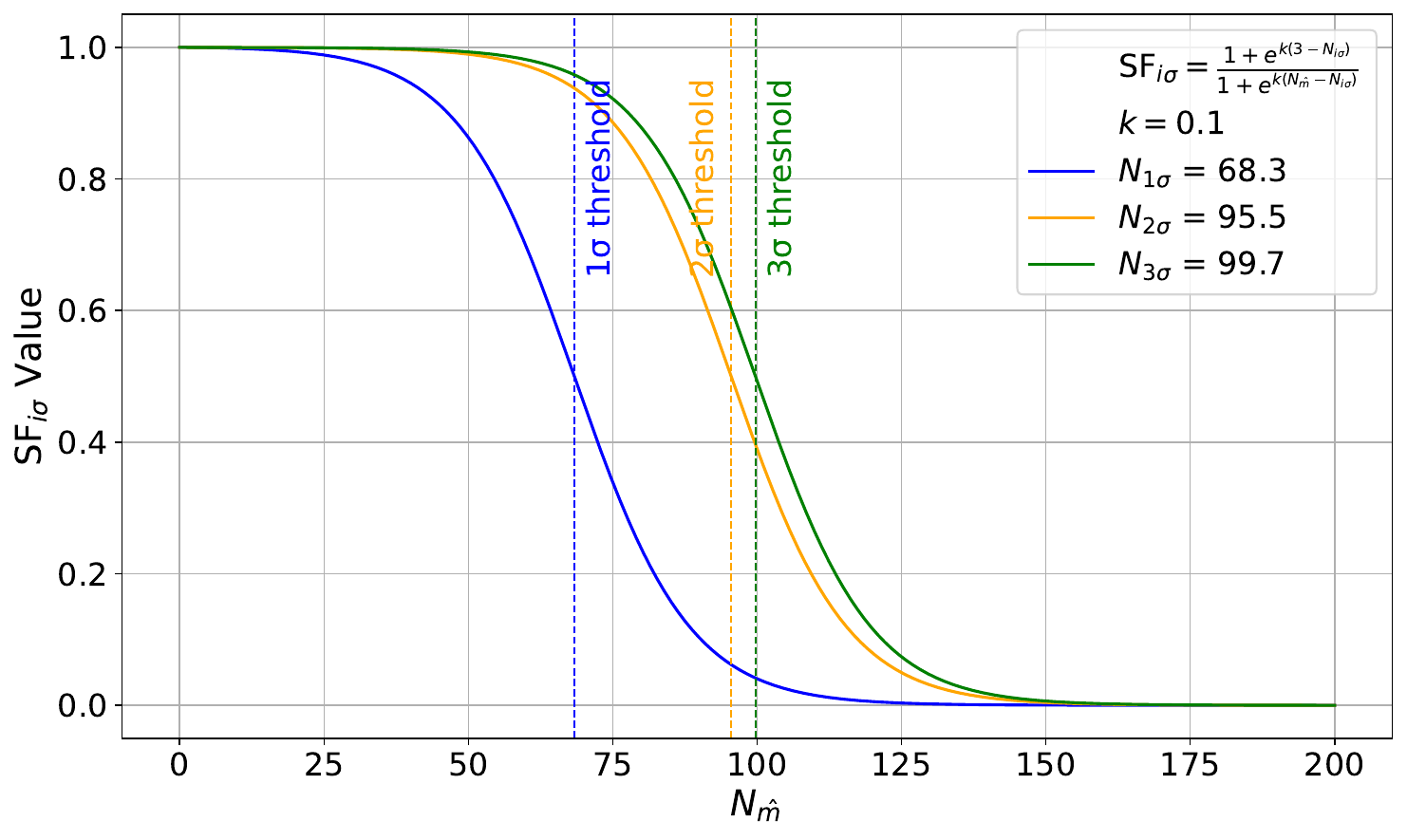}
  \caption{Simplicity Factor (SF) function curves with different vertex thresholds $N_{i\sigma}$. Each curve corresponds to a different threshold value ($N_{1\sigma}$, $N_{2\sigma}$ and $N_{3\sigma}$) with the decay rate $k=0.1$. }
  \label{fig:sf_curve}
\end{figure}

\begin{figure}[tb]
  \centering
  \includegraphics[width=0.8\columnwidth]{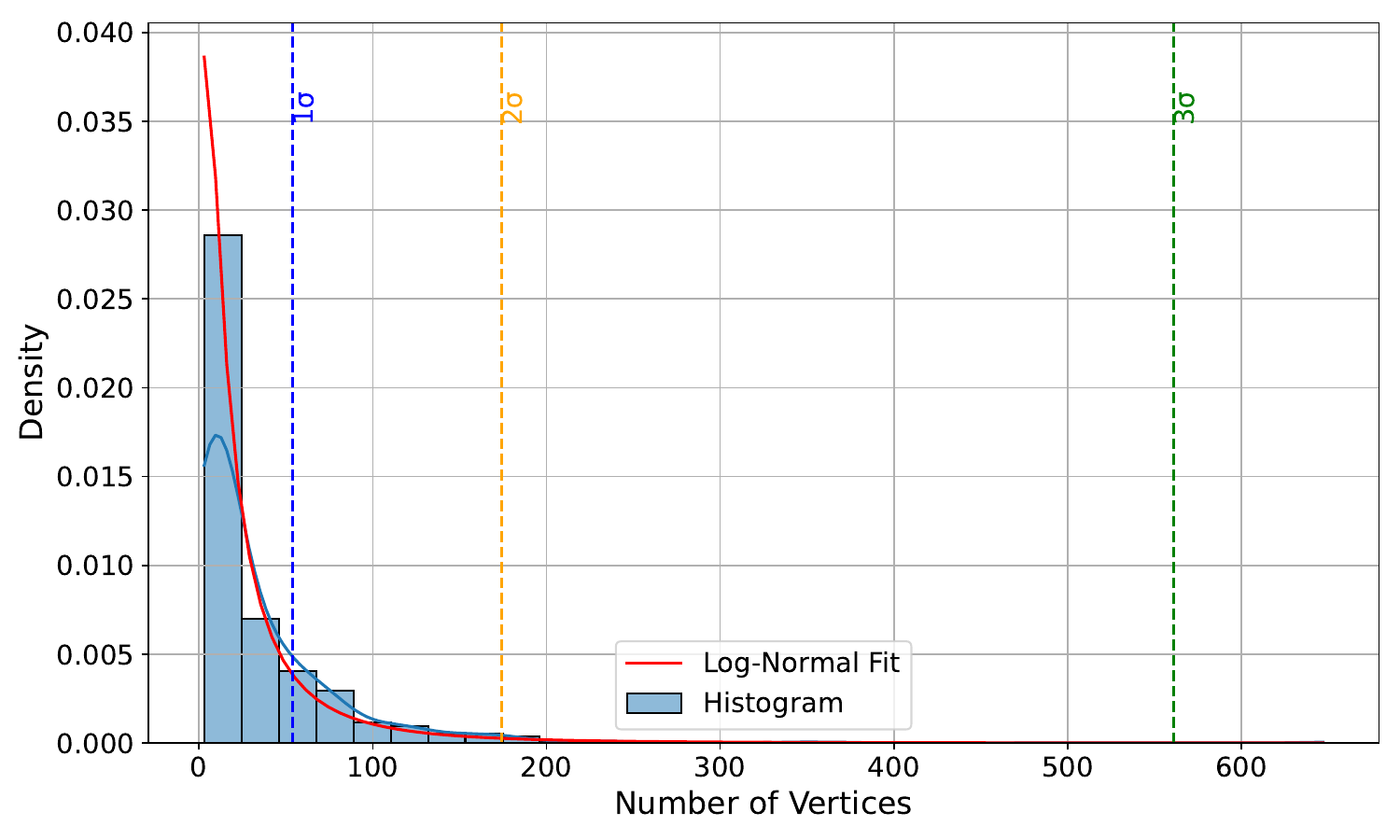}
  \caption{The statistical distribution of vertex counts in the Map2ImLas dataset, with a fitted log-normal curve (red). The upper bounds of the 1$\sigma$, 2$\sigma$, and 3$\sigma$ intervals are indicated by blue, yellow, and green dashed lines, respectively.}
  \label{fig:data_statistics}
\end{figure}

As illustrated in Figure \ref{fig:sf_curve}, SF is a bounded function ranging from 0 to 1. 
When the predicted polygon has the minimum number of vertices ($N_{\hat{m}}=3$), SF reaches its maximum value of 1, indicating no penalty for highly simplified shapes. 
As $N_{\hat{m}}$ increases, SF gradually decreases. 
The decline is slow at first, becomes steep around the vertex threshold $N_{i\sigma}$, and then flattens. 
As $N_{\hat{m}}\rightarrow\infty$, SF asymptotically approaches zero. 
This design provides a balanced and interpretable penalty function. 
It allows the model to moderately increase the number of vertices, particularly when $N_{\hat{m}} \ll N_{i\sigma}$, to better preserve the polygonal shape. 
Meanwhile, it imposes a sharp penalty around $N_{i\sigma}$, which represents the upper bound of vertex counts observed in most road polygons, to discourage unnecessary over-detailing. 
The flattened behavior beyond the threshold avoids strong penalties for predictions that slightly exceed the typical vertex range. 
SF is formally defined as:
\begin{equation}
    \text{SF}(N_{\hat{m}}) = \sum_{i=1}^{3}\text{SF}_{i\sigma} =  \sum_{i=1}^{3}\frac{(1+\mathrm{e}^{k\cdot(3-N_{i\sigma})})}{1 + \mathrm{e}^{k\cdot(N_{\hat{m}}-N_{i\sigma})}}
\end{equation}
where the exponent $k$ is the decay rate, controlling how steeply the $\rm SF$ value drops near $N_{i\sigma}$. 
Since $N_{i\sigma}$ acts as the upper bound of typical vertex counts, its choice directly affects the penalty behavior. 
A small value may over-penalize complex but reasonable predictions, while a large value may be too tolerant to over-complexity. 
To set an appropriate $N_{i\sigma}$, we analyze the statistical distribution of vertex counts in the Dutch road dataset, as shown in Figure \ref{fig:data_statistics}. 
Although the ground truth polygons are not strictly optimized for simplicity, their vertex count distribution still offers a meaningful reference. 
This is because the penalty function exhibits smooth and stable behavior around the chosen thresholds. 
Hence, using slightly larger values will not lead to abrupt changes or unreasonable penalization. 
We observe that the vertex counts of road polygons $N_m$ approximately follow a log-normal distribution:
\begin{equation}
f(N_m;\mu,\sigma)=\frac{1}{N_m\sigma\sqrt{2\pi}}\mathrm{e}^{-\frac{(\ln N_m - \mu)^2}{2\sigma^2}}
\end{equation}
where $\mu$ and $\sigma$ denote the log-mean and log-scale standard deviation. 
For a log-normal distribution, about 68.3\%, 95.5\%, and 99.7\% of the data fall within $1\sigma, 2\sigma,$ and $3\sigma$ ranges around the mean. 
We thus set $N_{1\sigma}, N_{2\sigma}, N_{3\sigma}$ to correspond to the 1$\sigma$, 2$\sigma$, and 3$\sigma$ intervals of the fitted distribution, respectively, and calculate the mean value of them as the final SF value. This multi-scale design ensures robustness across polygons of varying complexity. 

Notably, S-IoU enables reference-free evaluation of polygon simplicity during inference, as it only requires dataset-level statistics. 
This makes it especially useful in semi-supervised or annotation-limited settings where ground truth contours may be unavailable. 

\noindent\textbf{Polygon regularity and polygon connectivity.} 
Since road networks are typically interconnected with many branches, the correctness of their connectivity is crucial for large-scale topographic mapping. 
Moreover, roads often span long distances with smooth and regular outlines. 
Therefore, evaluating both the regularity and connectivity of the predicted road polygons is essential to fully reflect their geometric and topological quality. 

To assess polygon regularity, we first utilize the PoLiS metric \cite{avbelj2014metric}, which quantifies shape consistency between predicted and ground truth polygons by measuring the average distance from each vertex of one polygon to the boundary of the other, and vice versa. 
However, PoLiS only measures the overall shape consistency while neglecting boundary smoothness. 
To address this limitation, we propose a new metric, termed Smoothness Consistency Ratio (SCR). 
It is defined as the ratio between the number of inflection points, $N_\text{inflect}^\text{pred}$ and $N_\text{inflect}^\text{gt}$, in the predicted and ground truth polygons. 
\begin{equation}
    \text{SCR} = \frac{N_\text{inflect}^\text{pred}}{N_\text{inflect}^\text{gt}}
\end{equation}
An inflection point is defined as a vertex where the angle between two consecutive edges exceeds a predefined threshold. 
While road junctions and structural elements such as bus stops and parking areas may introduce inflection points, the overall number remains limited due to the inherently smooth nature of road outlines. 
Therefore, a large SCR often indicates jagged or noisy outlines, while a small SCR suggests over-smoothed predictions. 
An ideal SCR close to 1 implies consistent smoothness with the ground truth. 

To evaluate the topological connectivity of predicted road polygons, we adopt the Average Pah Length Similarity (APLS) metric \cite{van2018spacenet}. 
APLS is a graph-based metric. 
Let $a$ and $b$ denote a pair of nodes in the ground truth graph $G$, $\hat{a}$ and $\hat{b}$ denote their nearest counterparts in the predicted graph $\hat{G}$. 
APLS is formulated as:
\begin{equation}
    \text{APLS} = 1 - \frac{1}{n}\Sigma \min(\frac{|\mathrm{d}_G(a, b) - \mathrm{d}_{\hat{G}}(\hat{a}, \hat{b})|}{\mathrm{d}_G(a, b)}, 1)
\end{equation}
where $\mathrm{d}_G(a, b)$ and $\mathrm{d}_G(\hat{a}, \hat{b})$ represent the shortest path length between nodes $a$, $b$ on graph $G$ and nodes $\hat{a}$, $\hat{b}$ on graph $\hat{G}$, respectively. 
The APLS score ranges from 0 to 1, with higher values indicating better topological consistency. If a valid path exists between $a$ and $b$ in $G$ but not between $\hat{a}$ and $\hat{b}$ in $\hat{G}$, or vice versa, the APLS score is assigned zero. 

\begin{figure}[htbp]
  \centering
  \subfloat[Road polygon]{\includegraphics[width=0.3\textwidth]{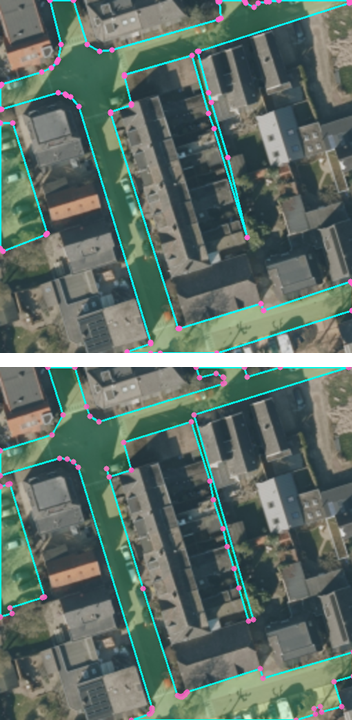}} \hspace{0.01\textwidth}
  \subfloat[Extracted road network]{\includegraphics[width=0.3\textwidth]{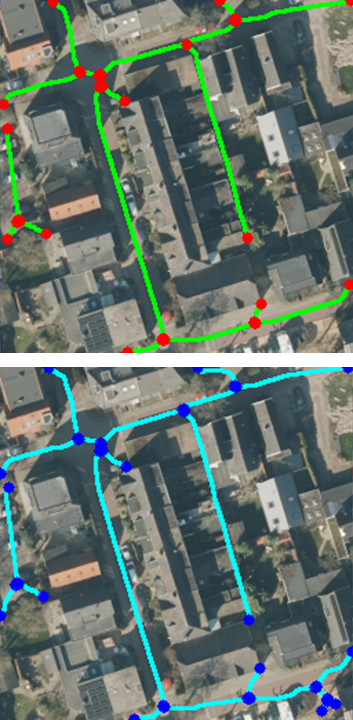}} \hspace{0.01\textwidth}
  \subfloat[Shortest path between nodes]{\includegraphics[width=0.3\textwidth]{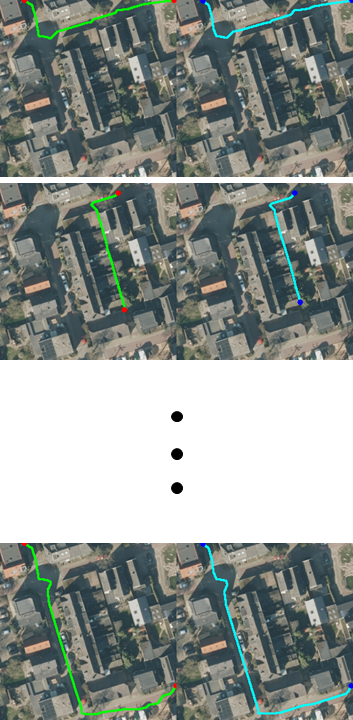}}
  \caption{Illustration of the APLS computation pipeline. Left: predicted (top) and ground truth road polygons (bottom). Middle: corresponding pseudo road networks extracted via skeletonization, where control nodes are highlighted. Right: shortest path examples between selected control node pairs on both graphs, used to calculate path length similarity.}
  \label{fig:apls}
\end{figure}

Since only road polygons are available, we first convert them into pseudo road networks via skeletonization, generating densely distributed nodes. We then identify endpoints and junctions as control nodes and compute the APLS metric accordingly, as illustrated in Figure \ref{fig:apls}.

\subsection{Experimental results and analysis}
\noindent\textbf{Results on the Deventer region.} 
Table \ref{tab:deventer_road} presents a quantitative comparison between our proposed method and existing methods on the Deventer region. 
Considering the similarity to our task, we first compare against state-of-the-art segmentation-based polygonal building outline extraction methods, including FFL \cite{girard2021polygonal} and HiSup \cite{xu2023hisup}. 
We further compare with representative approaches from road-related object extraction. 
Since road network extraction and road boundary extraction methods are not directly applicable to our task, we select road segmentation methods as baselines, as they can be adapted to polygonal road outline extraction through post-processing. 
Specifically, we apply the Douglas-Peucker algorithm to convert segmentation masks into vectorized polygons, with the simplification parameter set to 1 to balance vertex sparsity and IoU preservation, following \cite{xu2023hisup}. 
We include two categories of road segmentation methods in our comparison.
The first category consists of general segmentation baselines, including SegNet \cite{badrinarayanan2017segnet}, UNet \cite{ronneberger2015u}, and DeepLabV3+ \cite{chen2018encoder}. 
The second category includes representative methods specifically designed to enhance road connectivity, for which we select D-LinkNet\cite{zhou2018d} and CoANet~\cite{mei2021coanet}. 

As shown in Table \ref{tab:deventer_road}, our method demonstrates competitive performance across all evaluation metrics on the Deventer dataset. 
In particular, it demonstrates clear advantages in polygon regularity and polygon connectivity. 
For polygon regularity, it achieves the best SCR score, narrowing the gap to the ideal value of 1 by 0.22, and the lowest PoLiS value (2.60), indicating smoother and more geometrically aligned polygonal outlines compared to the ground truth. 
Notably, HiSup \cite{xu2023hisup}, despite leveraging additional geometric features and a hierarchical supervision scheme to enhance mask regularity, still underperforms our method in both metrics, particularly with a 4.19 point drop in SCR. 
For polygon connectivity, our method achieves the highest APLS score of 68.9, which is 3.3 points higher than the second-best method (HiSup). 
Even D-LinkNet~\cite{zhou2018d} and CoANet~\cite{mei2021coanet}, which are explicitly designed to enhance road connectivity, fall short, with APLS scores of 60.5 and 65.2, respectively. 
These results fully illustrate the effectiveness of our approach in preserving road topological correctness. 
Our method also excels in vertex-efficiency metrics. The C-IoU reaches 62.0, 5.7 points higher than the second-best method, and the N-ratio is closest to the ideal value (1.18), indicating better balance between segmentation accuracy and vertex efficiency. 
In terms of pixel-level segmentation performance, our method achieves the best IoU (75.0) and Boundary IoU (B-IoU) (67.4), reflecting accurate pixel-level coverage and boundary alignment. 
Finally, our proposed Simplicity-aware IoU (S-IoU), which encourages accurate segmentation with fewer vertices, achieves a score of 59.8, outperforming all baselines by a notable margin. This metric complements traditional evaluations and highlights the model’s capability to generate compact and high-quality polygons. 
\label{sec:experimantal_results}
\begin{table}[tb]
  \caption{Quantitative comparison with state-of-the-art models on the Deventer region. Best values are in boldface and second-best values are underlined. For N ratio and SCR, the reported values represent relative deviations from the ideal value of 1.}
  \label{tab:deventer_road}
  \centering
  \fontsize{8pt}{8pt}\selectfont
  \begin{tabular}{@{}l|ccccccccc@{}}
    \toprule
    Method & $\rm IoU$$\uparrow$ & $\rm B\text{-}IoU$$\uparrow$& $\rm C\text{-}IoU$$\uparrow$ & $\rm N \,ratio$$\rightarrow$1 & $\rm PoLiS$$\downarrow$ & $\rm S\text{-}IoU$$\uparrow$ & $\rm SCR$$\rightarrow$1 & $\rm APLS$$\uparrow$ \\
    \midrule
    SegNet \cite{badrinarayanan2017segnet} & 64.0 & 54.8 & 37.4 & +2.52 & 5.66 & 34.7 & +2.67 & 54.6 \\
    UNet \cite{ronneberger2015u} & 67.1 & 57.4 & 46.3 & +1.32 & 4.67 & 42.9 & +1.36 & 56.8 \\
    DeepLabV3+ \cite{chen2018encoder} & 70.3 & 60.3 & 47.5 & +1.49 & 4.48 & 44.4 & +1.54 & 56,4 \\
    D-LinkNet \cite{zhou2018d} & 71.7 & 63.0 & 54.5 & +0.68 & 3.27 & 51.4 & \underline{+0.73} & 60.5 \\
    CoANet \cite{mei2021coanet} & \underline{71.9} & 61.6 & 51.4 & +1.08 & 3.86 & 47.9 & +1.12 & 65.2 \\
    FFL \cite{girard2021polygonal} & 57.1 & 47.1 & 28.4 & +4.78 & 5.47 & 25.2 & +4.82 & 56.5\\
    HiSup \cite{xu2023hisup} & 71.5 & \underline{63.1} & \underline{56.3} & \underline{+0.24} & \underline{2.95} & \underline{57.5} & +4.41 & \underline{65.6} \\ \midrule 
    LDPoly (Ours) & \textbf{75.0} & \textbf{67.4} & \textbf{62.0} & \textbf{+0.18} & \textbf{2.60} & \textbf{59.8} & \textbf{+0.22} & \textbf{68.9} \\
  \bottomrule
  \end{tabular}
\begin{flushleft}
\footnotesize
\end{flushleft}
\end{table}

\begin{figure}[t!]
\begin{subfigure}{\linewidth}
\centering
\includegraphics[width=\linewidth]{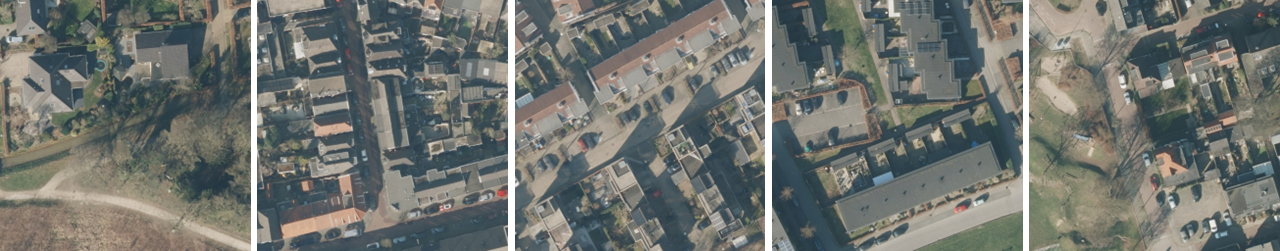}
\caption{Aerial image}
\end{subfigure}

\begin{subfigure}{\linewidth}
\centering
\includegraphics[width=\linewidth]{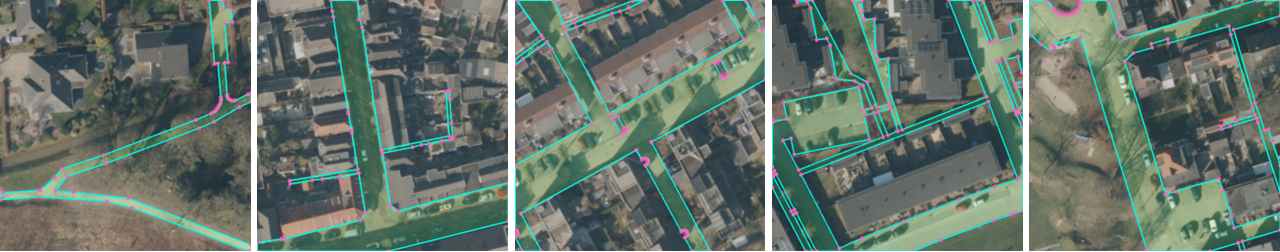}
\caption{Ground truth}
\end{subfigure}

\begin{subfigure}{\linewidth}
\centering
\includegraphics[width=\linewidth]{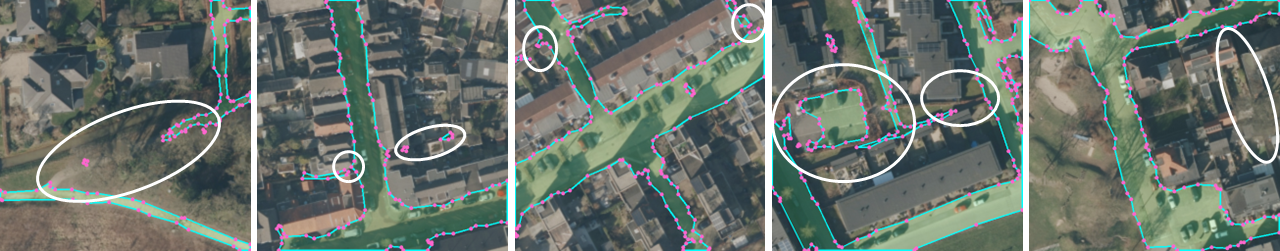}
\caption{UNet}
\end{subfigure}

\begin{subfigure}{\linewidth}
\centering
\includegraphics[width=\linewidth]{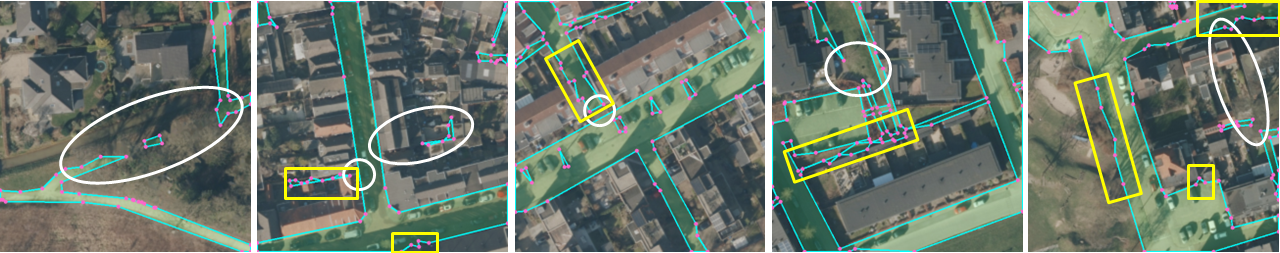}
\caption{HiSup}
\end{subfigure}

\begin{subfigure}{\linewidth}
\centering
\includegraphics[width=\linewidth]{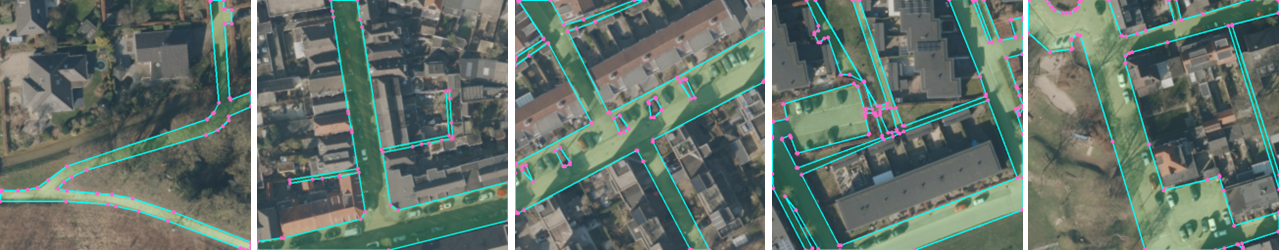}
\caption{LDPoly (ours)}
\end{subfigure}
\caption{Qualitative comparison of polygonal road outlines on the Deventer region. Top to bottom: aerial image, ground truth, UNet \cite{ronneberger2015u}, HiSup \cite{xu2023hisup}, and LDPoly (ours). White circles highlight road disconnectivity. Yellow boxes highlight irregular road outlines.}
\label{fig:deventer_road_qualitative_comparison}
\end{figure}

\begin{figure}[htbp]
  \centering
  \begin{minipage}[t]{0.495\linewidth}
    \centering
    \includegraphics[width=\linewidth]{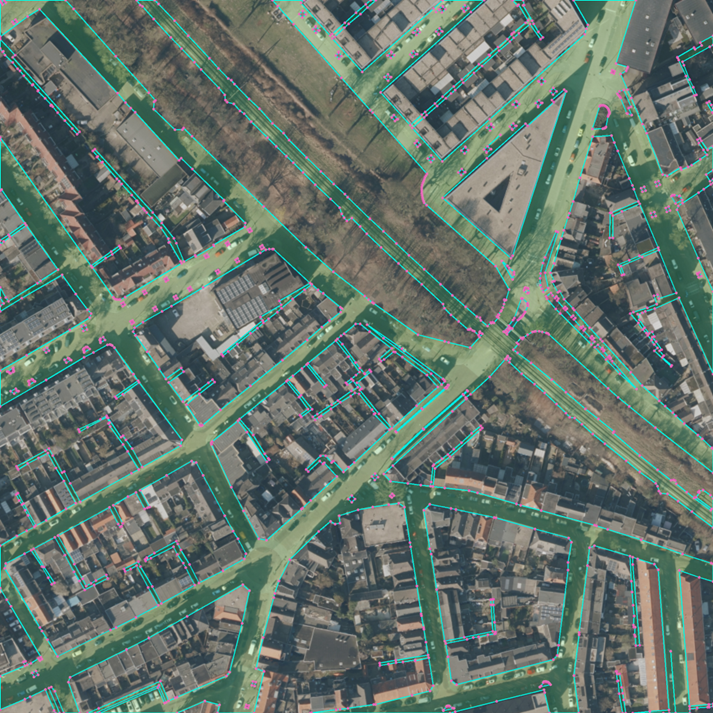}
    \subcaption{Ground truth}
  \end{minipage}%
  \hfill
  \begin{minipage}[t]{0.495\linewidth}
    \centering
    \includegraphics[width=\linewidth]{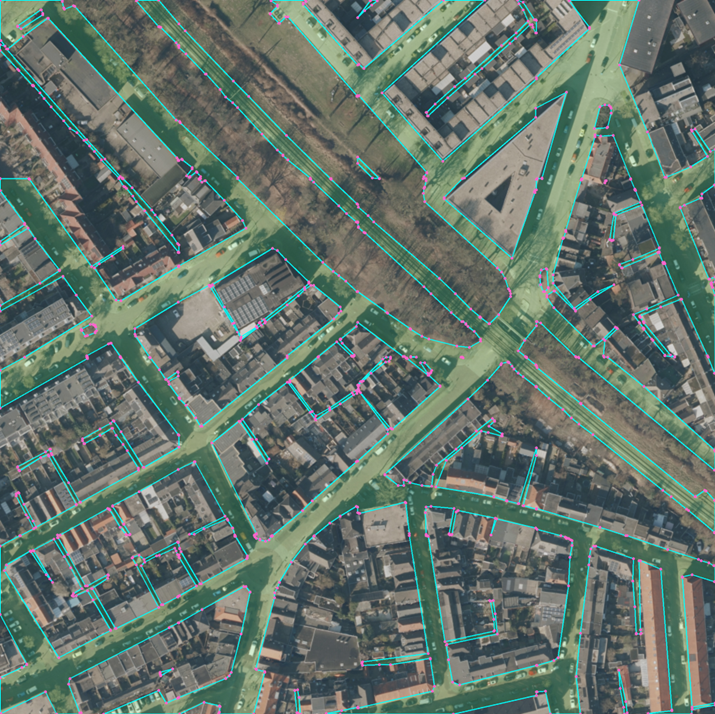}
    \subcaption{Ours}
  \end{minipage}
  \caption{Stitched visualization of large-scale polygonal road outlines in the Deventer region.}
  \label{fig:stitched_deventer_road}
\end{figure}

We further provide qualitative comparisons on several representative examples in Figure~\ref{fig:deventer_road_qualitative_comparison}. 
We compare our results with two representative baselines: UNet \cite{ronneberger2015u} and HiSup \cite{xu2023hisup}. 
HiSup is a state-of-the-art method that directly generates vectorized polygons and ranks second in most evaluation metrics. 
UNet, on the other hand, is a representative baseline which highlights common issues of segmentation pipelines followed by post-processing (Douglas-Peucker). 
As can be seen in the figure, UNet suffers from extremely irregular outlines with redundant vertices and jagged boundaries. 
Its road connectivity is also limited with fragmented segments and missed connections, as highlighted in the white circles. 
Compared to UNet, HiSup achieves much better outlines but struggles with jagged boundaries in visually obscure areas, such as thin alleyways and shadowed areas, as marked by the yellow boxes. 
Moreover, HiSup also exhibits notable road disconnectivity, as seen in the white circles. 
In comparison, benefiting from the strong generative and reasoning capabilities of the diffusion model, our method produces road polygons with noticeably better outline regularity and smoothness, as well as superior road connectivity. 
In addition, our predicted road polygons demonstrate significantly lower vertex redundancy compared to other methods. 
Our results also exhibit improved polygon simplicity compared to the ground truth. In particular, at curved road junctions, the underlying geometry is preserved with fewer vertices. 
Notably, our model remains effective in visually obscured conditions, such as shadow-covered dirt roads (first column) and thin alleyways (second, fourth and fifth columns). 
Figure \ref{fig:stitched_deventer_road} presents a stitched visualization by merging spatially adjacent patches of the test set into a large image, providing a more intuitive illustration of the overall boundary regularity and road connectivity of our results. 
To achieve this, we first stitch together the predicted masks of the 16 patches belonging to the same large image. We then aggregate the predicted vertices from these patches and adjust their coordinates based on their respective patch positions within the large image. Finally, polygonization is performed on the assembled large image to produce the large vectorized road polygon. 

\noindent\textbf{Cross-data test on the Enschede and Giethoorn regions.}
To assess the generalization ability of our model, we perform cross-data evaluations by directly applying the model trained on the Deventer region to two unseen regions. 

\begin{figure}[t!]
\begin{subfigure}{\linewidth}
\centering
\includegraphics[width=\linewidth]{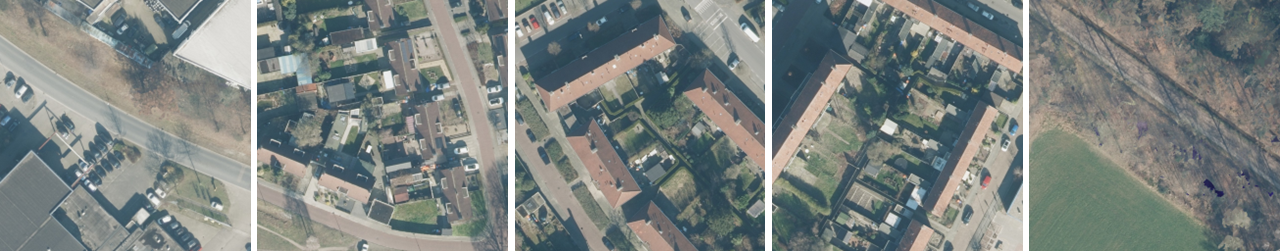}
\caption{Aerial image}
\end{subfigure}

\begin{subfigure}{\linewidth}
\centering
\includegraphics[width=\linewidth]{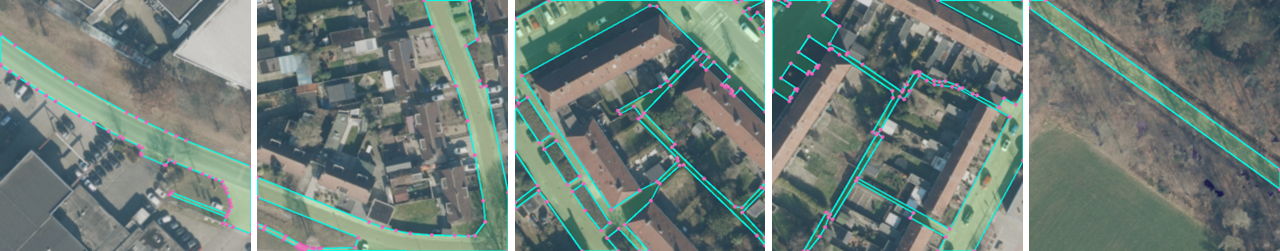}
\caption{Ground truth}
\end{subfigure}

\begin{subfigure}{\linewidth}
\centering
\includegraphics[width=\linewidth]{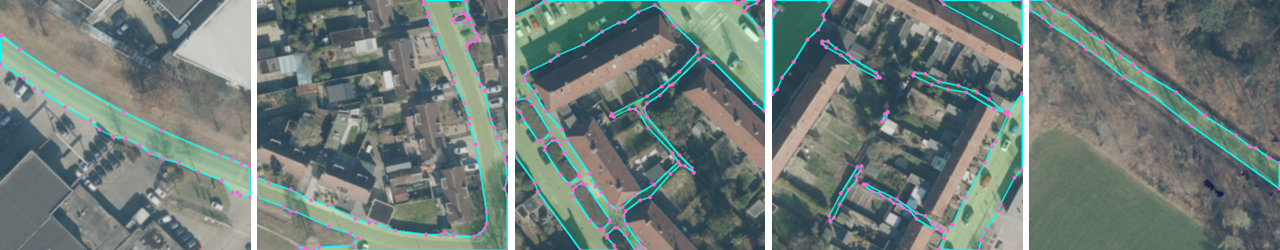}
\caption{DLinkNet}
\end{subfigure}

\begin{subfigure}{\linewidth}
\centering
\includegraphics[width=\linewidth]{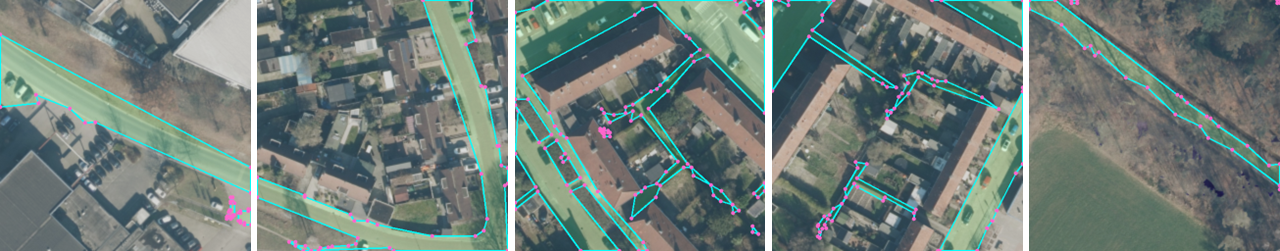}
\caption{HiSup}
\end{subfigure}

\begin{subfigure}{\linewidth}
\centering
\includegraphics[width=\linewidth]{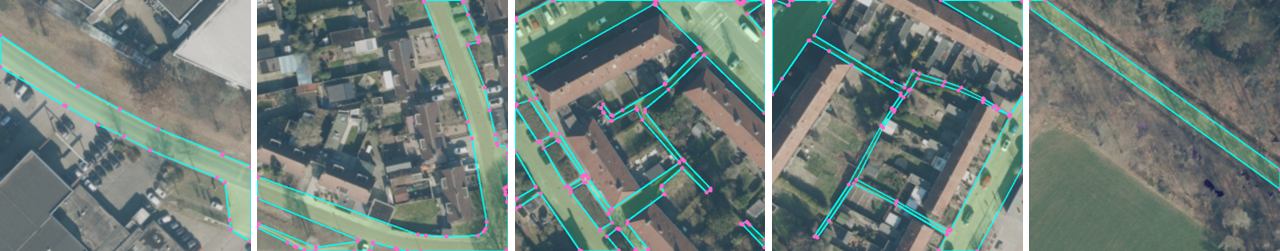}
\caption{LDPoly (ours)}
\end{subfigure}
\caption{Qualitative comparison of polygonal road outlines on the Enschede region. Top to bottom: aerial image, ground truth, DLinkNet \cite{zhou2018d}, HiSup~\cite{xu2023hisup}, and LDPoly (ours).}
\label{fig:enschede_road_qualitative_comparison}
\end{figure}

We first test on the Enschede region, a city located in the east of the Netherlands that generally exhibits urban road structures similar to Deventer. However, Enschede contains a larger number of industrial areas, where the presence of internal roads introduces additional annotation ambiguities. 
This setup enables us to assess the model's generalizability under both structurally consistent patterns and more challenging labeling conditions. 
As shown in Table \ref{tab:enschede_road}, our method achieves the best performance across most evaluation metrics. This includes significant improvements in polygon regularity (PoLiS, SCR, B-IoU), vertex efficiency (C-IoU, N-ratio), and polygon simplicity (S-IoU). 
These gains can be attributed to the strong generative capability of diffusion models in learning the underlying data distribution, which allows it to consistently maintain polygonal regularity and simplicity across varying scenes. 
Although our method does not achieve the highest IoU and APLS scores, it still ranks second in both. This slight gap is likely due to the annotation ambiguities in the Enschede dataset, particularly in industrial areas with internal road structures. 
Such inconsistencies pose challenges for generative models like ours that aim to capture the underlying data distribution. 
We further discuss this issue in Section~\ref{sec:discussion}. 
Figure~\ref{fig:enschede_road_qualitative_comparison} presents qualitative comparisons on the Enschede region. 
Since D-LinkNet \cite{zhou2018d} achieves the highest IoU and competitive APLS scores, we replace UNet with D-LinkNet for qualitative comparison in this region to ensure a more meaningful comparison. 
Note that HiSup remains included to ensure consistency with the Deventer evaluation. 
The results show that our method produces road polygons with overall lower vertex redundancy and more regular polygon outlines compared to other methods. 
Notably, in challenging areas such as thin alleyways (e.g., third and fourth examples) and visually obscured regions (e.g., fifth example), our model consistently outperforms other methods. 
In addition, the first example highlights the model’s ability to achieve high polygon simplicity. For a long, curved road boundary, our method preserves the underlying geometry using fewer but more effectively placed vertices. 
\begin{table}[tb]
  \caption{Quantitative comparison with state-of-the-art models on the Enschede region. Best values are in boldface and second-best values are underlined. For N ratio and SCR, the reported values represent relative deviations from the ideal value of 1.
  }
  \label{tab:enschede_road}
  \centering
  \fontsize{8pt}{8pt}\selectfont
  \begin{tabular}{@{}l|ccccccccc@{}}
    \toprule
    Method & $\rm IoU$$\uparrow$ & $\rm B\text{-}IoU$$\uparrow$& $\rm C\text{-}IoU$$\uparrow$ & $\rm N \,ratio$$\rightarrow$1 & $\rm PoLiS$$\downarrow$ & $\rm S\text{-}IoU$$\uparrow$ & $\rm SCR$$\rightarrow$1 & $\rm APLS$$\uparrow$1 \\
    \midrule
    SegNet \cite{badrinarayanan2017segnet} & 49.9 & 34.1 & 26.0 & +4.06 & 9.33 & 21.4 & +4.91 & 39.2 \\
    UNet \cite{ronneberger2015u} & 44.4 & 31.4 & 27.6 & +2.28 & 8.79 & 23.2 & +2.32 & 37.9 \\
    DeepLabV3+ \cite{chen2018encoder} & 63.1 & 46.6 & 40.7 & +2.02 & 6.62 & 35.8 & +2.05 & 47.7 \\
    D-LinkNet \cite{zhou2018d} & \textbf{69.6} & \underline{56.0} & 51.4 & +0.66 & 4.88 & 46.5 & +0.60 & 55.3 \\
    CoANet \cite{mei2021coanet} & 67.2 & 52.0 & 46.2 & +1.08 & 5.37 & 39.5 & +0.99 & \textbf{57.6} \\
    FFL \cite{girard2021polygonal} & 58.2 & 41.7 & 26.1 & +6.19 & 8.44 & 23.9 & +5.91 & 51.6 \\
    HiSup \cite{xu2023hisup} & 66.6 & 54.3 & \underline{51.7} & \underline{+0.43} & \underline{4.84} & \underline{52.7} & \underline{+0.48} & 55.4 \\ \midrule 
    Ours & \underline{69.0} & \textbf{56.8} & \textbf{55.0} & \textbf{+0.01} & \textbf{4.16} & \textbf{56.4} & \textbf{-0.04} & \underline{56.2} \\
  \bottomrule
  \end{tabular}
\begin{flushleft}
\footnotesize
\end{flushleft}
\end{table}

We then test our model on the Giethoorn region, a well-known rural tourist area in the Netherlands characterized by dense waterways, farmlands, and relatively sparse, branchless road structures. Unlike the urban layouts of Deventer and Enschede, Giethoorn presents a much simpler spatial pattern and homogeneous background, providing a distinct domain for evaluating cross-region generalization. 

\begin{figure}[t!]
\begin{subfigure}{\linewidth}
\centering
\includegraphics[width=\linewidth]{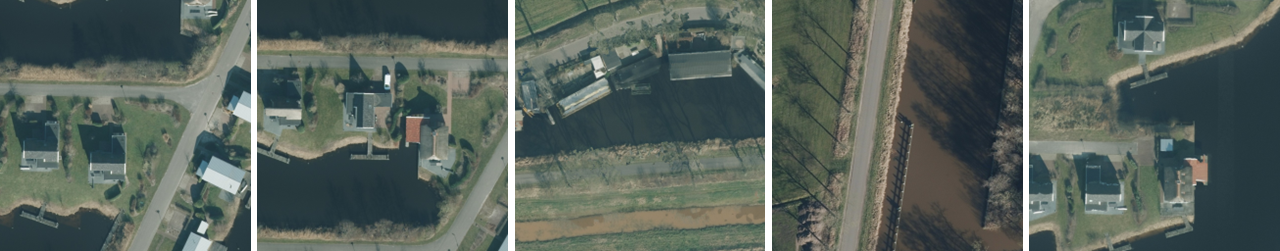}
\caption{Aerial image}
\end{subfigure}

\begin{subfigure}{\linewidth}
\centering
\includegraphics[width=\linewidth]{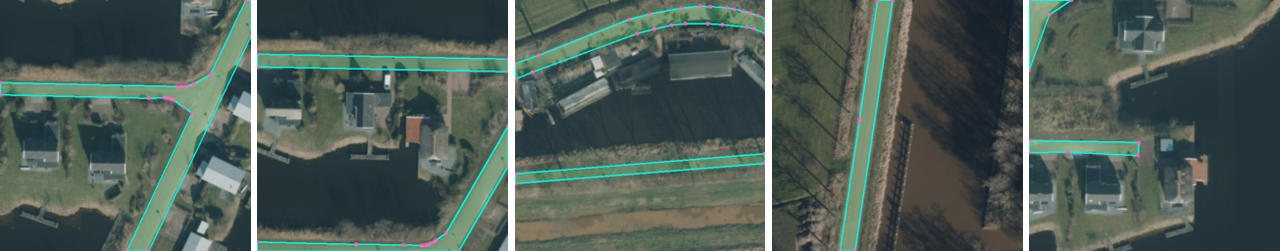}
\caption{Ground truth}
\end{subfigure}

\begin{subfigure}{\linewidth}
\centering
\includegraphics[width=\linewidth]{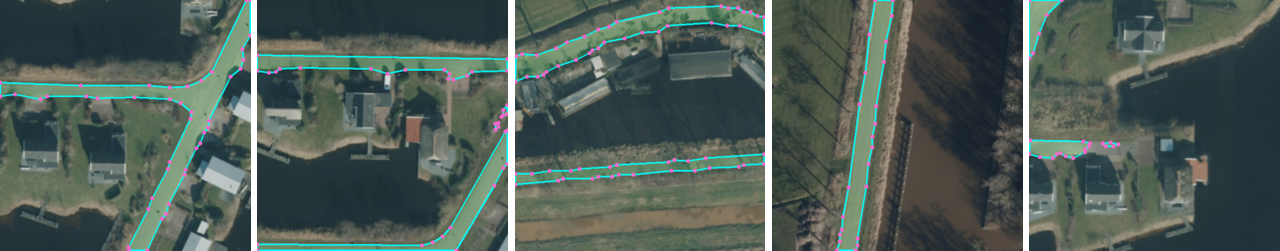}
\caption{DLinkNet}
\end{subfigure}

\begin{subfigure}{\linewidth}
\centering
\includegraphics[width=\linewidth]{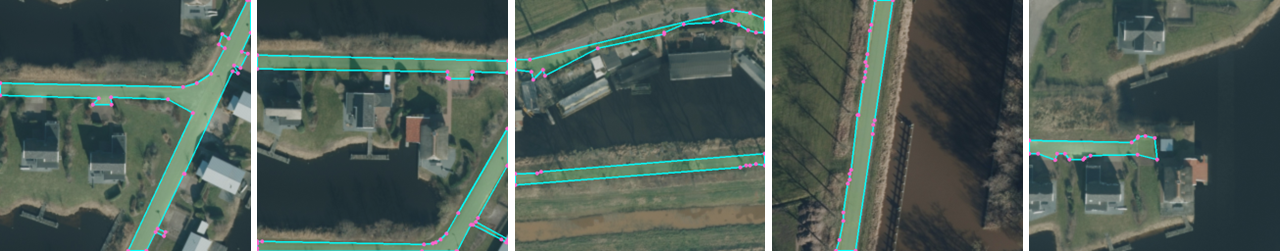}
\caption{HiSup}
\end{subfigure}

\begin{subfigure}{\linewidth}
\centering
\includegraphics[width=\linewidth]{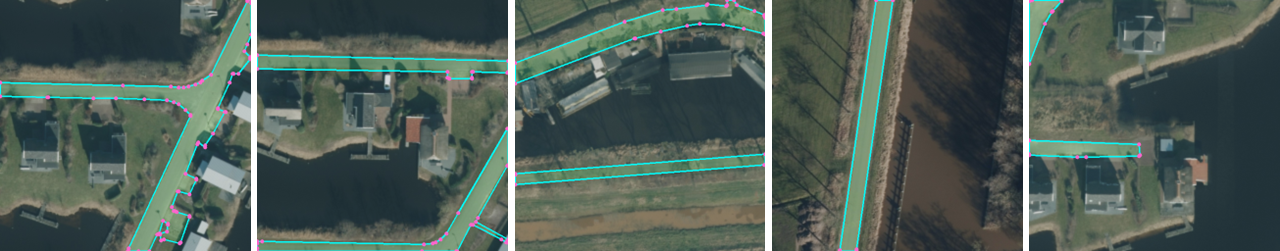}
\caption{LDPoly (ours)}
\end{subfigure}
\caption{Qualitative comparison of polygonal road outlines on the Giethoorn region. Top to bottom: aerial image, ground truth, DLinkNet \cite{zhou2018d}, HiSup~\cite{xu2023hisup}, and LDPoly (ours).}
\label{fig:geethorn_road_qualitative_comparison}
\end{figure}

As shown in Table \ref{tab:geethorn_road}, our method maintains strong overall performance. It achieves the best scores in key metrics such as C-IoU (55.0), S-IoU (62.4), and N-ratio (1.21), and ranks second in IoU (69.2), Boundary IoU (62.5), and APLS (71.1). 
Although our IoU, BIoU and PoLiS are slightly lower than D-LinkNet, this is primarily due to the difference in model behavior under domain shift. 
Giethoorn’s road annotations are relatively simplified, typically omitting minor paths and parking areas, which are emphasized in the Deventer dataset. 
This makes the dataset more favorable for discriminative models like D-LinkNet, which often struggle to preserve fine-grained structural details and instead focus on learning dominant patterns from the data, as evidenced by their limited performance in the Deventer region. 
In contrast, our generative model demonstrates a stronger capacity to produce more complex road polygons with detailed structures. 
However, this may lead to false positives (e.g., minor roads or private parking areas near main roads) that are practically reasonable in urban contexts but are not annotated in the rural setting, as illustrated in the first and second examples of Figure \ref{fig:geethorn_road_qualitative_comparison}. 
Additionally, HiSup slightly outperforms our method in SCR (+0.16 vs. +0.24), largely due to its use of parallel-edge merging, which reduces vertex redundancy but sometimes leads to over-smoothed boundaries, especially on long curved roads, as shown in the third example of Figure \ref{fig:geethorn_road_qualitative_comparison}. 
Despite these minor gaps, our method still delivers competitive road connectivity (APLS of 71.1 vs. CoANet’s 71.4) and outperforms all baselines in vertex efficiency and polygon simplicity. 
These results confirm our model’s robustness and its ability to generate compact yet accurate polygonal road representations across diverse spatial domains. 

\begin{table}[tb]
  \caption{Quantitative comparison with state-of-the-art models on the Giethoorn region. Best values are in boldface and second-best values are underlined. For N ratio and SCR, the reported values represent relative deviations from the ideal value of 1.
  }
  \label{tab:geethorn_road}
  \centering
  \fontsize{8pt}{8pt}\selectfont
  \begin{tabular}{@{}l|ccccccccc@{}}
    \toprule
    Method & $\rm IoU$$\uparrow$ & $\rm B\text{-}IoU$$\uparrow$& $\rm C\text{-}IoU$$\uparrow$ & $\rm N \,ratio$$\rightarrow$1 & $\rm PoLiS$$\downarrow$ & $\rm S\text{-}IoU$$\uparrow$ & $\rm SCR$$\rightarrow$1 & $\rm APLS$$\uparrow$1 \\
    \midrule
    SegNet \cite{badrinarayanan2017segnet} & 51.9 & 45.0 & 22.7 & +4.75 & 3.85 & 28.3 & +5.15 & 53.1 \\
    UNet \cite{ronneberger2015u} & 56.6 & 50.5 & 34.2 & +2.98 & 3.38 & 41.5 & +3.28 & 57.8 \\
    DeepLabV3+ \cite{chen2018encoder} & 64.7 & 57.1 & 43.0 & +1.46 & 2.57 & 54.5 & +1.62 & 61.0 \\
    D-LinkNet \cite{zhou2018d} & \textbf{71.9} & \textbf{65.6} & 52.6 & \underline{+0.89} & \textbf{2.01} & \underline{62.1} & +0.97 & 70.4 \\
    CoANet \cite{mei2021coanet} & 64.9 & 56.3 & 42.8 & +1.72 & 2.67 & 51.5 & +1.85 & \textbf{71.4} \\
    FFL \cite{girard2021polygonal} & 50.3 & 44.0 & 21.4 & +7.51 & 3.85 & 27.5 & +8.02 & 65.8 \\
    HiSup \cite{xu2023hisup} & 67.6 & 61.3 & \underline{52.8} & \textbf{+0.21} & \underline{2.21} & \underline{62.1} & \textbf{+0.16} & 67.6 \\ \midrule 
    Ours & \underline{69.2} & \underline{62.5} & \textbf{55.0} & \textbf{+0.21} & 2.28 & \textbf{62.4} & \underline{+0.24} & \underline{71.1} \\
  \bottomrule
  \end{tabular}
\begin{flushleft}
\footnotesize
\end{flushleft}
\end{table}

\subsection{Ablation study}
\label{sec:ablation}
We conduct several ablation studies to further validate the effectiveness of our proposed method. 
All experiments were conducted on the Deventer region under the same experimental setup as introduced in Section \ref{sec:experimantal_setup}. 

\noindent\textbf{The effectiveness of the diffusion-based generative scheme.} To fairly evaluate the advantage of a diffusion-based generative framework over a purely discriminative one, we perform an ablation study by removing the diffusion scheme from LDPoly, reducing the model to a denoising UNet operating in the latent space. 
This variant directly predicts road vertex heatmaps and road masks from aerial images without any iterative denoising process. 
We compare its performance with LDPoly to highlight the benefits brought by the diffusion scheme. 

\begin{figure}[t!]
\begin{subfigure}{\linewidth}
\centering
\includegraphics[width=\linewidth]{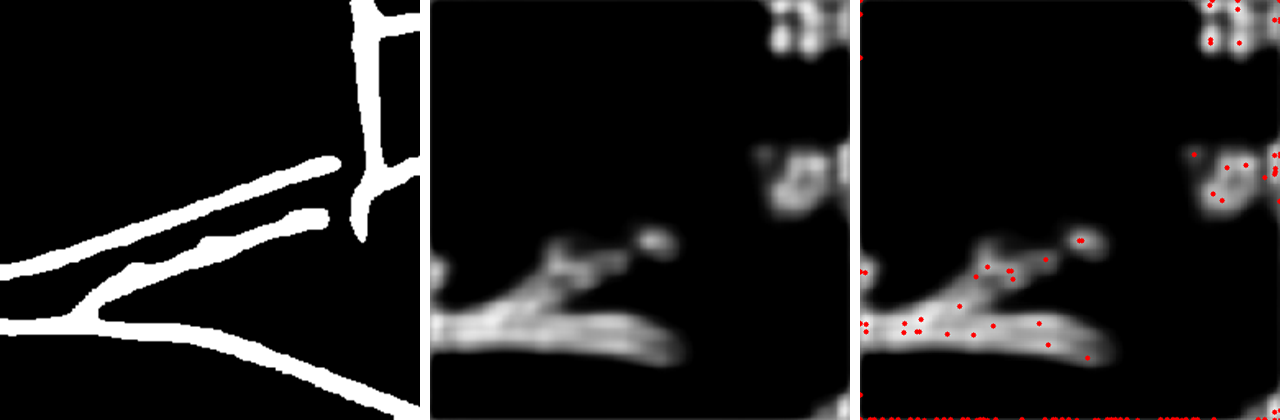}
\caption{w/o diffusion}
\end{subfigure}
\begin{subfigure}{\linewidth}
\centering
\includegraphics[width=\linewidth]{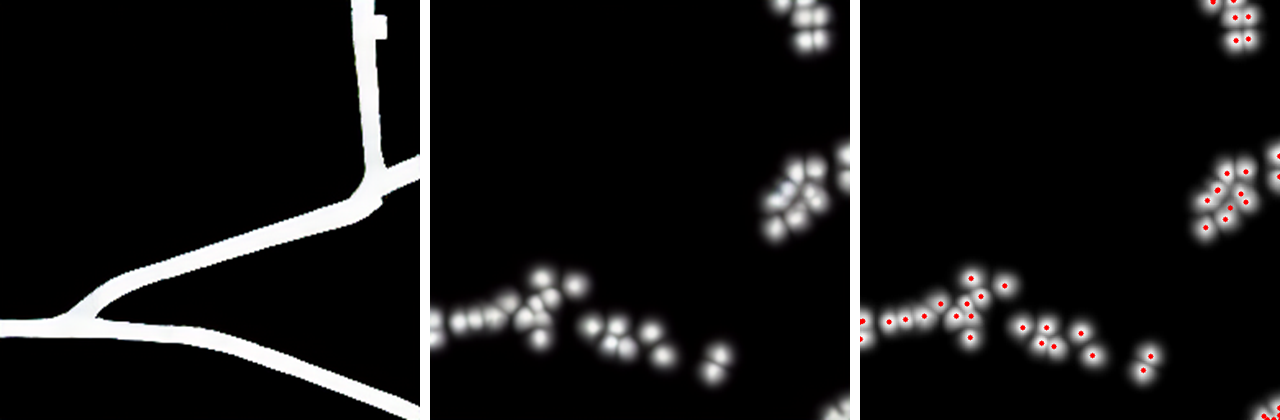}
\caption{w/ diffusion (ours)}
\end{subfigure}
\caption{An example comparing predicted road mask, vertex heatmap, and extracted vertex candidates between with and without diffusion. Extracted vertices are shown as red points on the heatmap.}
\label{fig:no_diffusion_vs_diffusion1}
\end{figure}

In Table \ref{tab:no_diffusion_vs_diffusion1}, we first compare the predicted road masks and vertices of the two models before polygonization. 
As shown in the table, LDPoly outperforms the non-diffusion baseline, with a 2.7 point improvement in IoU and a larger 4.5 point gain in B-IoU. 
This notable increase in B-IoU highlights the effectiveness of the diffusion scheme in generating more regular and accurate road polygons. 
As for vertex prediction, we apply non-maximum suppression (NMS) to extract vertex candidates from the predicted heatmaps and evaluate them against ground-truth vertices. A prediction is considered correct if it falls within 10 pixels of a ground-truth vertex. Under this criterion, LDPoly achieves significantly higher precision (79.8 vs. 26.0) and recall (75.4 vs. 70.6), highlighting the generative model’s superior capability to localize sparse and detail-sensitive vertices. 
Figure \ref{fig:no_diffusion_vs_diffusion1} presents a visual comparison of predicted masks, vertex heatmaps, and extracted vertex candidates. 
The baseline model without diffusion exhibits degraded boundary regularity, lower road connectivity, and highly blurred heatmaps with many false positives and missing details, highlighting the importance of the diffusion-based generative scheme in polygonal road outline extraction. 

\begin{figure}[t!]
\begin{subfigure}{\linewidth}
\centering
\includegraphics[width=\linewidth]{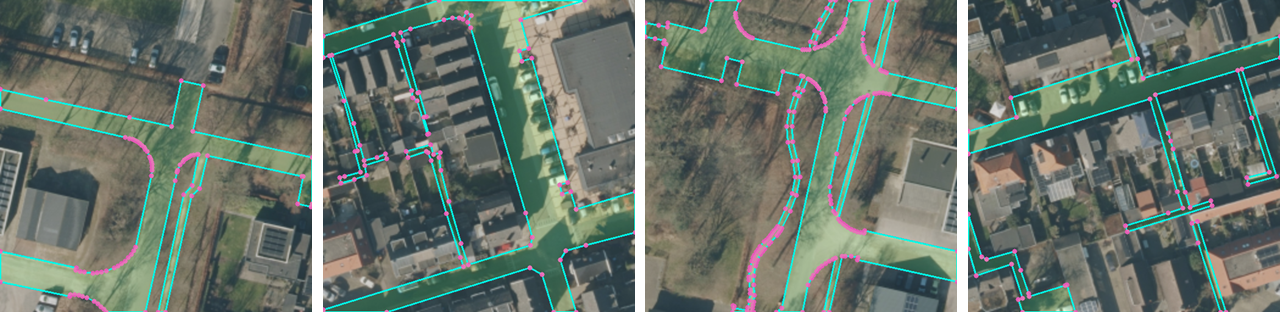}
\caption{ground truth}
\end{subfigure}
\begin{subfigure}{\linewidth}
\centering
\includegraphics[width=\linewidth]{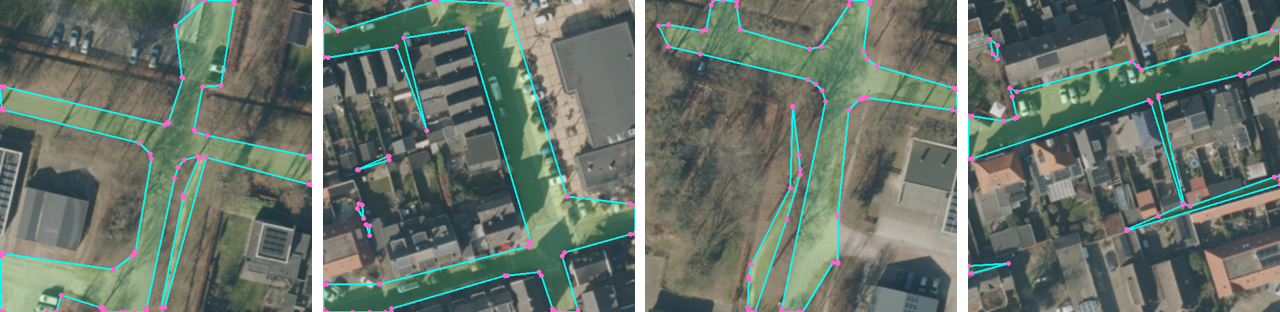}
\caption{w/o diffusion}
\end{subfigure}
\begin{subfigure}{\linewidth}
\centering
\includegraphics[width=\linewidth]{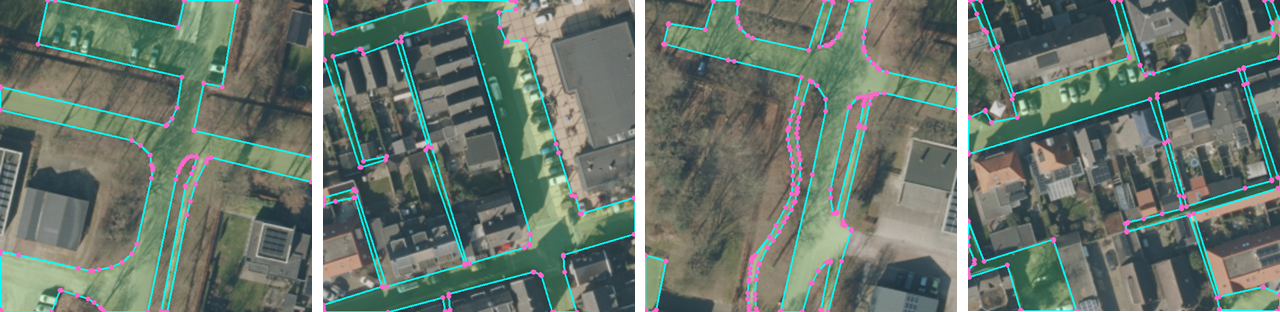}
\caption{w/ diffusion (ours)}
\end{subfigure}
\caption{Qualitative comparison of polygonal road outlines between with and without diffusion.}
\label{fig:no_diffusion_vs_diffusion2}
\end{figure}

Table \ref{tab:no_diffusion_vs_diffusion2} further shows evaluation results after polygonization. Due to the poor vertex predictions, the baseline model experiences significant drops across all metrics: IoU, B-IoU, C-IoU, PoLiS, and APLS. As illustrated in Figure \ref{fig:no_diffusion_vs_diffusion2}, the resulting polygons from the non-diffusion model are often fragmented and oversimplified. 
This is caused by disconnected road masks and missing vertices. 
The reduced number of predicted vertices slightly improves N-ratio, S-IoU, and SCR but at the cost of geometric and topological quality. 
These experiments provide sufficient and fair evidence that incorporating the diffusion-based generative scheme significantly improves the regularity and connectivity of road polygons, as well as enhances the generation of sparse vertex heatmaps that require strong sensitivity to local geometric details. 
\begin{table}[tb]
  \caption{Quantitative comparison of mask and vertex prediction between with and without diffusion.}
  \label{tab:no_diffusion_vs_diffusion1}
  \centering
  \fontsize{8pt}{8pt}\selectfont
  \begin{tabular}{@{}l|cc|cc@{}}
    \toprule
    Method & $\rm IoU$$\uparrow$ & $\rm B\text{-}IoU$$\uparrow$& Vertex Prec.$\uparrow$ & Vertex Rec.$\uparrow$ \\
    \midrule
    w/o Diffusion & 74.0 & 65.3 & 26.0 & 70.6 \\
    w/ Diffusion (ours) & \textbf{76.7} & \textbf{69.8} & \textbf{79.8} & \textbf{75.4} \\
  \bottomrule
  \end{tabular}
\end{table}

\begin{table}[tb]
  \caption{Quantitative comparison of polygonal road outlines between with and without diffusion.}
  \label{tab:no_diffusion_vs_diffusion2}
  \centering
  \fontsize{8pt}{8pt}\selectfont
  \begin{tabular}{@{}l|cccccccc@{}}
    \toprule
    Method & $\rm IoU$$\uparrow$ & $\rm B\text{-}IoU$$\uparrow$& $\rm C\text{-}IoU$$\uparrow$ & $\rm N \,ratio$$\rightarrow$1 & $\rm PoLiS$$\downarrow$ & $\rm S\text{-}IoU$$\uparrow$ & $\rm SCR$$\rightarrow$1 & $\rm APLS$$\uparrow$ \\
    \midrule
    w/o Diffusion & 66.2 & 55.7 & 50.7 & \textbf{+0.17} & 3.34 & 57.1 & \textbf{+0.18} & 59.5 \\
    w/ Diffusion (ours) & \textbf{75.0} & \textbf{67.4} & \textbf{62.0} & +0.18 & \textbf{2.60} & \textbf{59.8} & +0.22 & \textbf{68.9} \\
  \bottomrule
  \end{tabular}
\end{table}

\noindent\textbf{The effectiveness of the proposed channel-embedded fusion module. }
The proposed channel-embedded fusion module (CEFM) is designed to facilitate efficient feature interaction between the two denoising targets. 
To validate its effectiveness, we conduct an ablation study by replacing CEFM with a simple concatenation of the latent vectors of the vertex heatmap and road mask along the last channel dimension. 
As shown in Table \ref{tab:CEFM}, this simple fusion strategy leads to a performance drop in most metrics, particularly in IoU, B-IoU, C-IoU, and PoLiS. 
This highlights the importance of CEFM in enabling effective feature interaction between the vertex heatmap and road mask, thereby leading to better shape alignment with the ground truth road polygons. 
While slightly better results are observed in S-IoU, N-ratio, and SCR, this can be attributed to degraded vertex extraction. 
As shown in Table \ref{tab:CEFM2}, although the IoU and BIoU scores of using concatenation are comparable to those of CEFM, the vertex recall is notably lower, indicating more miss detections. 
This results in oversimplified polygons with fewer vertices but lower geometric fidelity. 
\begin{table}[tb]
  \caption{Quantitative comparison of polygonal road outlines between concatenation and CEFM.}
  \label{tab:CEFM}
  \centering
  \fontsize{8pt}{8pt}\selectfont
  \begin{tabular}{@{}l|cccccccc@{}}
    \toprule
    Method & $\rm IoU$$\uparrow$ & $\rm B\text{-}IoU$$\uparrow$& $\rm C\text{-}IoU$$\uparrow$ & $\rm N \,ratio$$\rightarrow$1 & $\rm PoLiS$$\downarrow$ & $\rm S\text{-}IoU$$\uparrow$ & $\rm SCR$$\rightarrow$1 & $\rm APLS$$\uparrow$ \\
    \midrule
    concat & 73.5 & 65.8 & 60.2 & \textbf{+0.11} & 2.82 & \textbf{59.9} & \textbf{+0.15} & 67.7  \\
    CEFM & \textbf{75.0} & \textbf{67.4} & \textbf{62.0} & +0.18 & \textbf{2.60} & 59.8 & +0.22 & \textbf{68.9} \\
  \bottomrule
  \end{tabular}
\end{table}

\begin{table}[tb]
  \caption{Quantitative comparison of mask and vertex prediction between concatenation and CEFM.}
  \label{tab:CEFM2}
  \centering
  \fontsize{8pt}{8pt}\selectfont
  \begin{tabular}{@{}l|cc|cc@{}}
    \toprule
    Method & $\rm IoU$$\uparrow$ & $\rm B\text{-}IoU$$\uparrow$& Vertex Prec.$\uparrow$ & Vertex Rec.$\uparrow$ \\
    \midrule
    concat & 75.3 & 68.4 & \textbf{79.8} & 72.4 \\
    CEFM & \textbf{76.7} & \textbf{69.8} & \textbf{79.8} & \textbf{75.4} \\
  \bottomrule
  \end{tabular}
\end{table}

\noindent\textbf{The effectiveness of the proposed polygonization method.} To fully evaluate the effectiveness of our proposed polygonization method, we compare it with two baselines: the classic Douglas-Peucker (DP) algorithm \cite{douglas1973algorithms} and the polygonization method used in HiSup \cite{xu2023hisup}. 
Additionally, we include an experiment to assess the impact of our Retaining Missing Inflection Points strategy by comparing results before and after its application. 

\begin{figure}[t!]
  \centering
  \begin{minipage}[t]{0.33\linewidth}
    \centering
    \includegraphics[width=\linewidth]{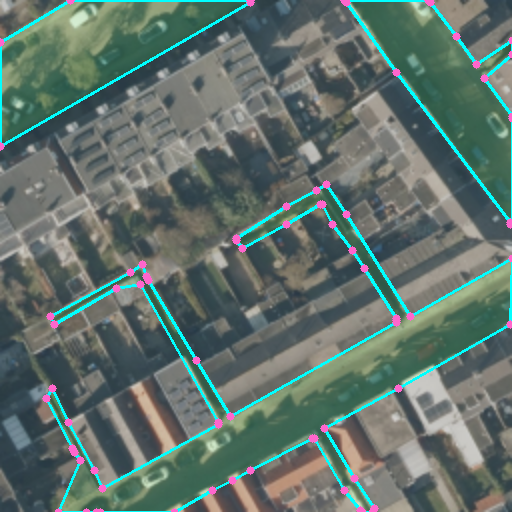}
    \subcaption{Our poly\textsuperscript{1}}
  \end{minipage}%
\hspace{0.02\linewidth}
  \begin{minipage}[t]{0.33\linewidth}
    \centering
    \includegraphics[width=\linewidth]{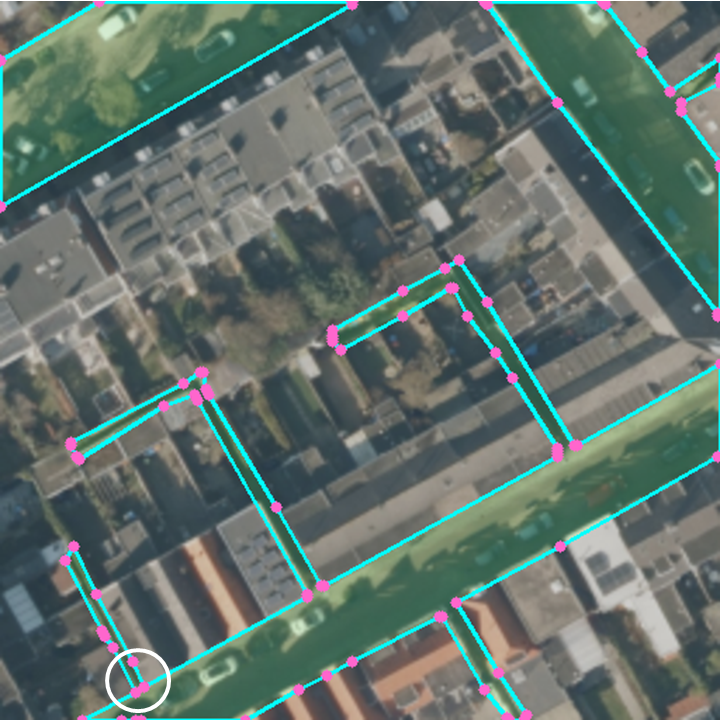}
    \subcaption{Our poly\textsuperscript{2}}
  \end{minipage}
  \caption{Comparison of polygonal road outlines before (Our poly\textsuperscript{1}) and after (Our poly\textsuperscript{2}) retaining missing inflection points. The retained inflection point is highlighted in the white circle. }
  \label{fig:retain_inflection_point}
\end{figure}

As shown in Table \ref{tab:polygonization}, although the DP algorithm achieves the highest IoU and BIoU scores, it introduces redundant vertices, leading to poor vertex efficiency and polygon simplicity, as evidenced by the C-IoU, N ratio and S-IoU metrics. 
Furthermore, without the guidance of extracted vertices, it produces irregular road polygon boundaries, leading to a significantly lower SCR score. 
By incorporating extracted vertices, the HiSup polygonization method addresses these issues. 
However, it suffers from a notable drop in IoU and BIoU, suggesting limitations in fully utilizing vertex information to reconstruct accurate vectorized road polygons. 
In contrast, our method exhibits minimal performance drop after polygonization, maintaining high pixel-level accuracy while achieving superior vertex efficiency and simplicity. 
Furthermore, with the inclusion of missing inflection points, our approach yields additional gains in IoU and BIoU with better preservation of local geometry, as illustrated in
Figure~\ref{fig:retain_inflection_point}. 
\begin{table}[tb]
  \caption{Quantitative comparison of different polygonization methods. Our poly\textsuperscript{1} denotes the version without retaining missing inflection points, and our poly\textsuperscript{2} includes this step.}
  \label{tab:polygonization}
  \centering
  \fontsize{8pt}{8pt}\selectfont
  \begin{tabular}{@{}l|cccccccc@{}}
    \toprule
    Method & $\rm IoU$$\uparrow$ & $\rm B\text{-}IoU$$\uparrow$& $\rm C\text{-}IoU$$\uparrow$ & $\rm N \,ratio$$\rightarrow$1 & $\rm PoLiS$$\downarrow$ & $\rm S\text{-}IoU$$\uparrow$ & $\rm SCR$$\rightarrow$1 & $\rm APLS$$\uparrow$ \\
    \midrule
    DP poly \cite{douglas1973algorithms} & \textbf{75.6} & \textbf{68.3} & 44.9 & +2.23 & \textbf{2.52} & 42.1 & +3.37 & 68.1  \\
    HiSup poly \cite{xu2023hisup} & 72.3 & 64.5 & 59.0 & +0.20 & 2.72 & 58.6 & +0.33 & 66.9 \\
    Our poly\textsuperscript{1} & 74.1 & 66.4 & 60.3 & \textbf{+0.05} & \underline{2.60} & \textbf{62.4} & \textbf{+0.09} & \underline{68.8} \\
    Our poly\textsuperscript{2} & \underline{75.0} & \underline{67.4} & \textbf{62.0} & \underline{+0.18} & \underline{2.60} & \underline{59.8} & \underline{+0.22} & \textbf{68.9} \\
  \bottomrule
  \end{tabular}
\end{table}

\section{Discussion}
\label{sec:discussion}
Despite the great performance on the Dutch road datasets, LDPoly still exhibits some limitations. 
These limitations mainly stem from the inherent challenges of the data and the characteristics of generative modeling. 

\begin{figure}[tb]
  \centering
  \begin{subfigure}[t]{0.32\textwidth}
    \centering
    \includegraphics[width=\linewidth]{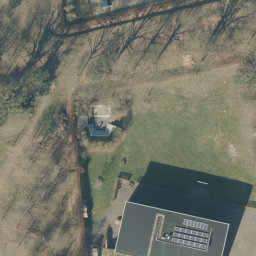}
    \caption{Aerial image}
  \end{subfigure}
  \hfill
  \begin{subfigure}[t]{0.32\textwidth}
    \centering
    \includegraphics[width=\linewidth]{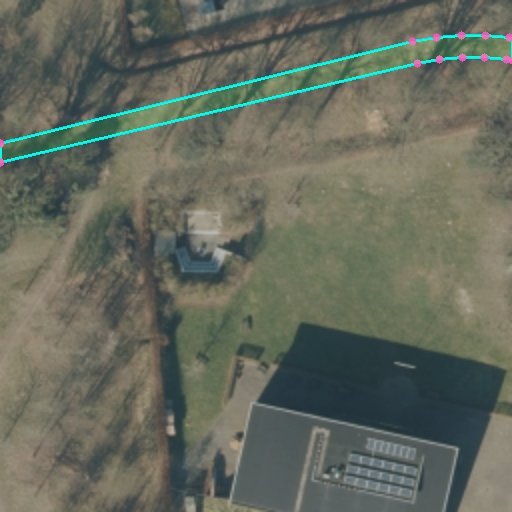}
    \caption{Ground truth}
  \end{subfigure}
  \hfill
  \begin{subfigure}[t]{0.32\textwidth}
    \centering
    \includegraphics[width=\linewidth]{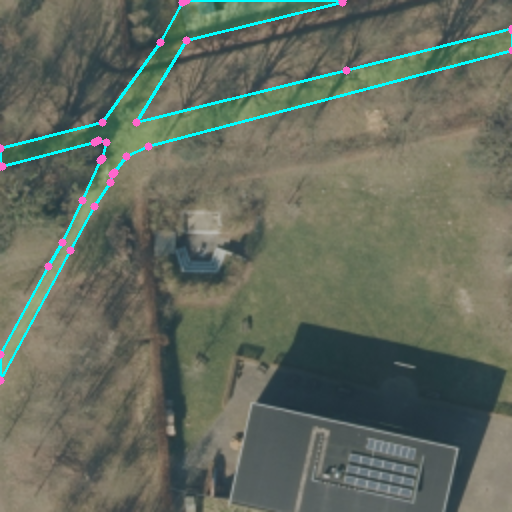}
    \caption{Ours}
  \end{subfigure}
  \caption{Example of false positive predictions in the Deventer region.}
  \label{fig:false_positive}
\end{figure}

One of the primary challenges is the tendency of our model to identify road-like structures that are not explicitly annotated in the ground truth. These false positives often occur in regions such as informal roads covered by dense tree shadows, dirt roads with unclear visual boundaries, or narrow alleyways that are visually obscured. 
While such predictions reflect LDPoly’s strong reasoning and generative capabilities, they may lead to undesirable false detections in practice. 
Figure \ref{fig:false_positive} shows an example of such false positives in the Deventer region. 

\begin{figure}[t!]
    \centering
    \includegraphics[width=1.0\linewidth]{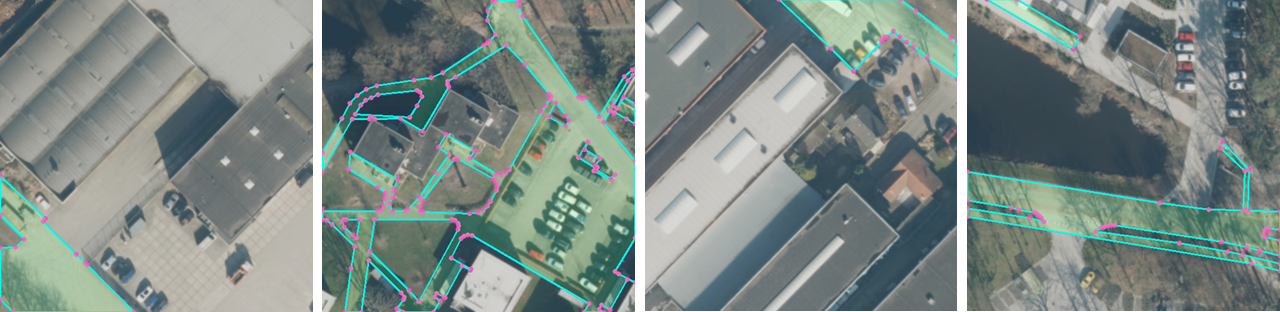}
    \caption{Examples of inconsistent annotations in the Enschede region.}
    \label{fig:annotation_issues}
\end{figure}

Another major limitation arises from inconsistent or incomplete annotations in the training data. In many instances, internal roads within factories or campuses are annotated inconsistently. 
Similarly, minor branches connecting main roads to private or restricted areas, such as parking lots or private properties, are often only partially labeled or completely omitted. 
We also observe clear cases of missing annotations in otherwise obvious road regions. 
For example, in the first example of Figure~\ref{fig:no_diffusion_vs_diffusion2}, the parking area at the top of the image is not considered as part of the road polygon, while the parking area at the bottom left is partially labeled as road and partially not. In the second example, the parking area to the right of the main road is fully annotated as road, whereas the adjacent ground surface, which lacks clear separation from the parking area, is not labeled as road. In the third example, two parking areas in the upper left corner are nearly invisible from the aerial image. In the fourth example, the ground in the lower right corner is not annotated as road, while a similar ground area in the lower left corner is labeled as part of the road polygon. 
We further show some examples of inconsistent annotations in the Enschede region in Figure \ref{fig:annotation_issues}. 
These inconsistencies pose significant challenges for generative models like LDPoly. 
Unlike discriminative models that may implicitly learn to ignore ambiguous signals, generative models are more sensitive to inconsistencies in training data, as they are trained to learn structured data distributions. 
This issue is particularly evident in the cross-domain evaluation on the Enschede dataset, where the presence of ambiguous internal roads contributes to a noticeable drop in IoU performance, despite our model’s strong results in other metrics. 

To mitigate these limitations, one potential solution is to incorporate strategies that explicitly enhance model robustness against label noise and visual ambiguity, such as uncertainty modeling. 
Another direction is to refine existing datasets by improving annotation consistency, particularly for ambiguous or partially labeled regions. 

\begin{figure}[tb]
  \centering
  \begin{subfigure}[t]{0.32\textwidth}
    \centering
    \includegraphics[width=\linewidth]{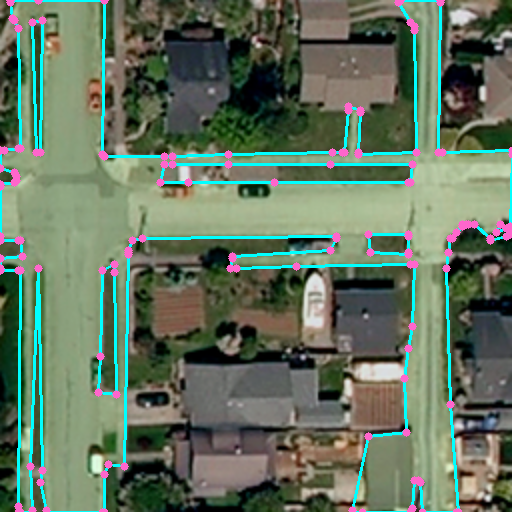}
    \caption{Bellingham}
  \end{subfigure}
  \hfill
  \begin{subfigure}[t]{0.32\textwidth}
    \centering
    \includegraphics[width=\linewidth]{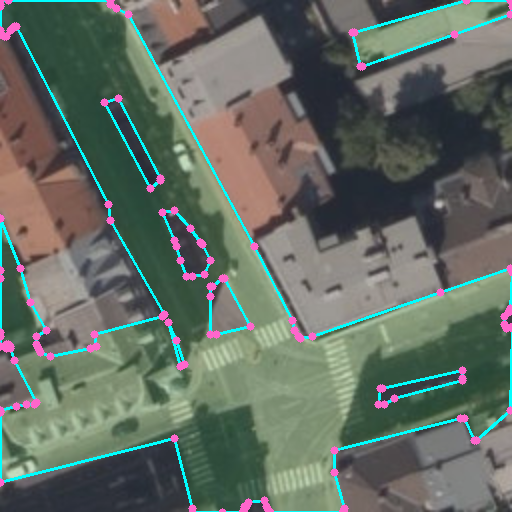}
    \caption{Innsbruck}
  \end{subfigure}
  \hfill
  \begin{subfigure}[t]{0.32\textwidth}
    \centering
    \includegraphics[width=\linewidth]{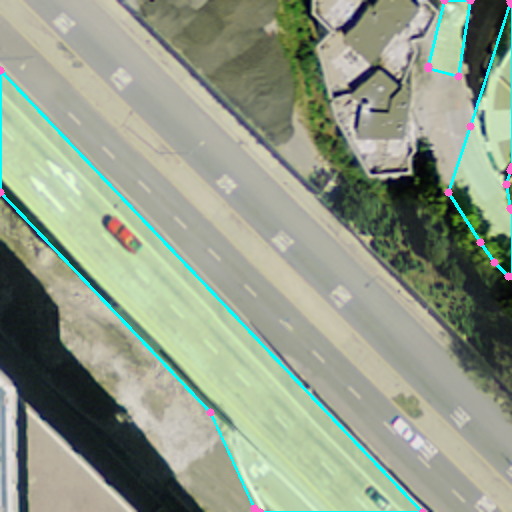}
    \caption{San Francisco}
  \end{subfigure}
  \caption{Example of cross-domain predictions in Bellingham, Innsbruck and San Francisco.}
  \label{fig:inria}
\end{figure}

To further evaluate the generalization capability of our model beyond the Dutch regions, we conducted an exploratory test on a few representative images from three regions of the INRIA building dataset \cite{maggiori2017dataset}: Bellingham, Innsbruck, and San Francisco. 
As shown in Figure \ref{fig:inria}, these regions represent decreasing levels of similarity to the road pattern of Deventer, which is our chosen training area in the Dutch road dataset. 
Bellingham, with its low urban density and narrow roads, closely resembles our training data. 
Innsbruck features wider roads, tall buildings, and ground shadows, which are uncommon in Deventer. 
San Francisco presents an even greater domain shift with the presence of elevated highways, completely absent in our training data. 
Without any fine-tuning, the model trained on Deventer generalizes well to Bellingham, indicating its capability to capture underlying urban road patterns. 
However, performance degrades in Innsbruck and San Francisco. 
These results highlight that while our model generalizes to visually similar environments, substantial domain differences still pose challenges, and future work could benefit from incorporating more diverse regional data to enhance cross-domain robustness.

Despite the aforementioned challenges, LDPoly consistently demonstrates the effectiveness of diffusion models in the task of polygonal road outline extraction. 
In particular, the model excels in generating geometrically regular and vertex-efficient polygon representations, while also preserving road connectivity. 
These results underscore the unique advantages of generative modeling in handling complex topological structures. 

\section{Conclusion}
In this paper, we propose a novel dual-latent diffusion model for polygonal road outline extraction, a critical task in large-scale topographic mapping. 
To harness the strong generative and reasoning capabilities of diffusion models, we design a dual denoising scheme along with a channel-embedded fusion module, enabling the simultaneous and effective denoising of vertex heatmaps and road masks. 
We also propose a polygonization algorithm to produce accurate and vectorized road outlines. 
To establish a benchmark for this task, we conduct a comprehensive quantitative evaluation of seven baseline models on a new dataset, facilitating fair comparison for future research. 
We also propose two novel evaluation metrics to better assess the boundary regularity and vertex efficiency of polygonal road outlines. 
Our work demonstrates the effectiveness of the diffusion-based generative model in generating polygonal outlines with superior regularity and topological correctness, compared to traditional discriminative models. 
We hope this work can offer new insights and serve as a foundation for future studies in topographic mapping and polygonal object outline extraction.

\section{Declaration of competing interest}
The authors declare that they have no known competing financial interests or personal relationships that could have appeared to influence the work reported in this paper.

\section{Acknowledgement}
This publication is part of the project ``Learning from old maps to create new ones", with project number 19206 of the Open Technology Programme which is financed by the Dutch Research Council (NWO). 
We would also like to thank Dr. Sander Oude Elberink and Geethanjali Anjanappa for providing access to the dataset used in this study, which is currently under review for publication.



 \bibliographystyle{elsarticle-num} 





\end{document}